\documentclass[oribibl]{llncs}

\usepackage[parfill]{parskip}    
\usepackage[version=0.96]{pgf}
\usepackage{tikz}
\usetikzlibrary{arrows,matrix,graphs,shapes,snakes,automata,backgrounds,petri,positioning,decorations.pathmorphing}
\usepackage[latin1]{inputenc}
\usepackage{verbatim}
\usepackage{amsmath}
\usepackage{amssymb}
\usepackage{latexsym}
\usepackage{proof}
\usepackage{array}
\usepackage{calc}
\usepackage{mathdots}
\usepackage{stmaryrd}
\usepackage{chicago}
\usepackage[english]{babel}
\usepackage{extarrows}

\newlength{\bre}
\newcolumntype{C}{@{}>{\rule{0mm}{\bre}}p{\bre}<{}@{}}

\newcommand{\editout}[1]{}

\newcommand{\ldl}{\mathbin{\backslash}}
\newcommand{\ldr}{\mathbin{/}}
\newcommand{\lpr}{\mathbin{\otimes}}
\newcommand{\gpr}{\mathbin{\oplus}}
\newcommand{\gdl}{\mathbin{\obslash}}
\newcommand{\gdr}{\mathbin{\oslash}}

\newcommand{\sempair}[2]{\langle #1,#2\rangle}

\newcommand{\mutilde}{\comu}

\newcommand{\labfrm}[2]{#1 \mathbin{:} #2}
\newcommand{\backsl}{\mathbin{\backslash}}

\newcommand{\apsnode}[2]{\overset{#1}{\underset{#2^{\rule{0pt}{1.2ex}}}{\centerdot}}}
\newcommand{\apsnodec}[1]{\underset{#1^{\rule{0pt}{1.2ex}}}{\centerdot}}

\newcommand{\apsnodeh}[1]{\overset{#1}{\centerdot}}
\newcommand{\apsnodei}{\centerdot_{\rule{0pt}{1.5ex}}}

\tikzset{pas/.style={fill=gray!60}, 
act/.style={fill=gray!30},
main/.style={draw,fill=white},
ctx/.style={rounded rectangle,minimum size=7mm},
val/.style={rectangle,minimum size=7mm},
cmd/.style={chamfered rectangle,draw,fill=white},
tns/.style={circle,minimum size=5mm,draw,fill=white},
par/.style={circle,minimum size=5mm,draw,fill=black}, 
minipar/.style={circle,minimum size=2.5mm,draw,fill=black}, 
pn/.style={rounded corners, rectangle,fill=blue!30,draw,minimum size=15mm},
medpn/.style={rounded corners, rectangle,fill=blue!30,draw,minimum size=20mm},
 bigpn/.style={rounded corners, rectangle,fill=blue!30,draw,minimum size=25mm}}

\tikzset{every picture/.style={scale=.9, transform shape, node distance=7mm}}
\newcommand{\lolli}{\multimap}

\newcommand{\arrow}[3]{#1:#2\longrightarrow #3}

\newcommand{\overa}[1]{\triangleright #1}
\newcommand{\undera}[1]{\triangleleft #1}
\newcommand{\overai}[1]{\triangleright^{-1} #1}
\newcommand{\underai}[1]{\triangleleft^{-1} #1}

\newcommand{\oslasha}[1]{\blacktriangleright #1}
\newcommand{\obslasha}[1]{\blacktriangleleft #1}
\newcommand{\oslashai}[1]{\blacktriangleright^{-1} #1}
\newcommand{\obslashai}[1]{\blacktriangleleft^{-1} #1}

\newcommand{\CPSlex}[1]{#1^{\ell}}

\newcommand{\W}[1]{\textsf{#1}}

\newcommand{\nd}[2]{#1 \vdash #2}
\newcommand{\Nd}[2]{\framebox[1.3\width]{\rule[-5pt]{0pt}{15pt}$#1$} \vdash #2}
\newcommand{\nD}[2]{#1 \vdash \framebox[1.3\width]{\rule[-5pt]{0pt}{15pt}$#2$}}

\newcommand{\arr}[2]{#1 \rightarrow #2}

\newcommand{\seq}[2]{#1\Rightarrow #2}

\newcommand{\Ra}{\rightarrow}

\newcommand{\bs}{\backslash}
\newcommand{\comu}{\widetilde{\mu}}

\newcommand{\Zip}[1]{\langle #1 \rangle}
\newcommand{\cmdL}[2]{\langle #1 \upharpoonleft #2 \rangle}
\newcommand{\cmdR}[2]{\langle #1 \upharpoonright #2 \rangle}

\newcommand{\Coco}[1]{\widetilde{#1}}
\newcommand{\LG}{\textbf{LG}}
\newcommand{\LaMu}{\mbox{$\overline{\lambda}\mu\comu$}}

 \newcommand{\leftleftharpoons}{%
  \mathbin{\ooalign{\hfil\raisebox{1pt}{$\leftharpoonup$}\hfil\cr\hfil
  \raisebox{-1pt}{$\leftharpoondown$}\hfil\crcr}}} 

 \newcommand{\rightrightharpoons}{%
  \mathbin{\ooalign{\hfil\raisebox{1pt}{$\rightharpoonup$}\hfil\cr\hfil
  \raisebox{-1pt}{$\rightharpoondown$}\hfil\crcr}}}

\newcommand{\otimesS}{\cdot\otimes\cdot}
\newcommand{\slashS}{\cdot\slash\cdot}
\newcommand{\bsS}{\cdot\bs\cdot}
\newcommand{\oplusS}{\cdot\oplus\cdot}
\newcommand{\oslashS}{\cdot\oslash\cdot}
\newcommand{\obslashS}{\cdot\obslash\cdot}
\newcommand{\Struct}[1]{\mathop{\cdot}#1\mathop{\cdot}}
\newcommand{\CPS}[1]{\lceil#1\rceil}

\begin{document}

\mainmatter              
\pagestyle{headings}  
\title{Proof nets for the Lambek-Grishin calculus}
\titlerunning{Proof nets for LG}  
%

\author{Michael Moortgat\inst{1} \and 
Richard Moot\inst{2}
\thanks{Draft of a chapter in E.~Grefenstette, C.~Heunen, and M.~Sadrzadeh (eds.)
`Compositional methods in Physics and Linguistics', OUP, to appear.
We thank Arno Bastenhof for helpful comments on an earlier version.
}
}
\authorrunning{Moortgat and Moot}   
%
\tocauthor{Michael Moortgat, Richard Moot}
\institute{Utrecht Institute of Linguistics OTS\\
Trans 10\\
3512 JK Utrecht, Netherlands\\
\email{M.J.Moortgat@uu.nl},\\ 
\and
CNRS, Universit\'e de Bordeaux\\
LaBRI, 351 cours de la Lib\'eration\\
33400 Talence, France\\
\email{moot@labri.fr}
}

\maketitle

\begin{abstract}
Grishin's generalization of Lambek's Syntactic Calculus combines
a non-commutative multiplicative conjunction and its residuals
(product, left and right division) with a dual family: multiplicative
disjunction, right and left difference. Interaction between these two
families takes the form of linear distributivity principles. 
We study proof nets for \LG\ and the correspondence between these nets and
unfocused and focused versions of its sequent calculus.
\end{abstract}

\section{Background, motivation}\label{intro}
In his two seminal papers \cite{lambek58,lambek61}, Jim Lambek
introduced the `parsing as deduction' method in linguistics: the traditional
parts of speech (noun, verb, adverb, determiner, etc) are replaced by 
logical formulas --- types if one takes the computational view;
the judgement whether an expression is well-formed is the outcome of
a process of logical deduction, or, reading formulas as types,
a computation in the type calculus.
\begin{equation}\label{timeflies}
\begin{array}{cccccccccc}
np & \otimes & (np\bs s) & \otimes & (((np\bs s)\bs (np\bs s))/np) & \otimes & (np/n) & \otimes & n & \Ra s\\
\W{time} & & \W{flies} & & \W{like} & & \W{an} & &  \W{arrow} & \\
\end{array}
\end{equation}
What is the precise nature of grammatical composition, the $\otimes$ operation in
the example above? The '58 and '61 papers present two views on this: in the '58 paper,
types are assigned to \emph{strings} of words, in the '61 paper, they are assigned
to \emph{phrases}, bracketed strings, with a grouping into constituents. The
Syntactic Calculus, under the latter view, is extremely simple. The derivability
relation between types is given by the preorder laws (\ref{preorder}) and the residuation
principles of (\ref{residuation}).
\begin{equation}
\label{preorder}
A\Ra A\qquad;\qquad\textrm{from $A\Ra B$ and $B\Ra C$ infer $A\Ra C$}
\end{equation}
\begin{equation}
\label{residuation}
A\Ra C/B\quad\textrm{iff}\quad A\otimes B\Ra C\quad\textrm{iff}\quad B \Ra A\bs C
\end{equation}
To obtain the '58 view, one adds the \emph{non-logical axioms} of (\ref{assoc}),
attributing associativity properties to the $\otimes$ operation.
\begin{equation}
\label{assoc}
(A\otimes B)\otimes C \Ra A\otimes(B\otimes C)
\qquad;\qquad
A\otimes(B\otimes C) \Ra (A\otimes B)\otimes C
\end{equation}
The Syntactic Calculus in its two incarnations --- the basic system \textbf{NL}
given by (\ref{preorder}) and (\ref{residuation}) and the associative variant \textbf{L}
which adds the postulates of (\ref{assoc}) --- recognizes only context-free
languages. It is well known that to capture the dependencies that occur in
natural languages, one needs expressivity beyond context-free. Here are
some characteristic patterns from formal language theory that can be
seen as suitable idealizations of phenomena that occur in the wild.
\begin{equation}\label{mcs}
\begin{array}{rl}
\textrm{copying:} &  \{w^{2} \mid w\in \{a,b\}^+ \}\\
\textrm{counting dependencies:} &  \{a^{n}b^{n}c^{n} \mid n>0 \}\\
\textrm{crossed dependencies:} &  \{a^{n}b^{m}c^{n}d^{m} \mid n,m>0\}\\
\end{array}
\end{equation}
In the tradition of extended rewriting systems, there is a large
group of grammar formalisms that handle these and related patterns gracefully:
Tree Adjoining Grammars, Linear Indexed Grammars, Combinatory Categorial
Grammars, Minimalist Grammars, Multiple Context Free Grammars, \ldots \cite{kallmeyer}.
Also in the Lambek tradition, extended type-logical systems have been proposed
with expressive power beyond context-free: multimodal grammars \cite{morr:type94,moor:mult95},
discontinuous calculi \cite{morrillea07}, etc. These extensions, as well as the original
Lambek systems, respect an ``intuitionistic'' restriction: in a sequent
presentation, derivability is seen as
a relation between (a structured configuration of) hypotheses $A_1,\ldots,A_n$ and
a \emph{single} conclusion $B$. In a paper antedating Linear Logic by a couple of years, \citeN{gris:onag83}
proposes a generalization of the Lambek calculus which removes this intuitionistic
restriction. Linguistic application of Grishin's ideas is fairly recent. In the
present paper, we study the system presented
in \cite{jfak60}, which we'll refer to as \LG.
\subsection{Dual residuation principles, linear distributivities}
In \LG\ 
 the inventory of type-forming operations
is doubled: in addition to the familiar operators $\otimes,\bs,/$ (product, left and right division),
we find a dual family $\oplus, \oslash,\obslash$: coproduct, right and left difference.
\begin{equation}
\label{vocabulary}
\begin{array}{lclcl}
A,B & ::= & p \mid & \qquad & \textrm{atoms: $s$, $np$, \ldots}\\
    & & A\otimes B \mid B\bs A \mid A/B \mid & \qquad &\textrm{product, left vs right division}\\
    & & A\oplus B \mid A\oslash B \mid B\obslash A & \qquad &\textrm{coproduct, right vs left difference}
\end{array}
\end{equation}
Some clarification about the notation: we follow \cite{lamb:from93} in writing $\oplus$ for the coproduct, which is
a multiplicative operation, like $\otimes$.  We read $B\bs A$ as `$B$ under $A$', $A/B$ as `$A$ over $B$',
$B\obslash A$ as `$B$ from $A$' and $A\oslash B$ as `$A$ less $B$'. For the difference operations, then,
the quantity that is subtracted is under the circled (back)slash, just as we have the denominator
under the (back)slash in the case of left and right division types. In a
formulas-as-types spirit, we will feel free to refer to the division operations as implications,
and to the difference operations as co-implications.
\paragraph*{Dual residuation principles} The most basic version of \LG\ is the symmetric
generalization of \textbf{NL}, which means that to (\ref{preorder}) and (\ref{residuation})
we add the dual residuation principles of (\ref{dualresiduation}).
\begin{equation}
\label{dualresiduation}
B\obslash C\Ra A\quad\textrm{iff}\quad C\Ra B\oplus A\quad\textrm{iff}\quad C\oslash A \Ra B
\end{equation}
To get a feeling for the consequences of the preorder laws (\ref{preorder}) and
the (dual) residuation principles (\ref{residuation}) and (\ref{dualresiduation}),
here are some characteristic theorems and derived rules of inference.
First, the \emph{compositions} of the product and division operations, and of
the co-product and difference operation give rise to the expanding and contracting
patterns of (\ref{coapplication}). The rows here are related by a left-right symmetry;
the columns by arrow reversal.
\begin{equation}
\label{coapplication}
\begin{array}{c@{\qquad}c}
A\otimes(A\bs B)\Ra B \Ra  A\bs(A\otimes B) &  (B/A)\otimes A\Ra  B\Ra  (B\otimes A)/A\\
(B\oplus A)\oslash A\Ra  B\Ra  (B\oslash A)\oplus A &  A\obslash(A\oplus B)\Ra  B\Ra  A\oplus(A\obslash B)\\
\end{array}
\end{equation}
Secondly, one can show that the type-forming operations have the monotonicity
properties summarized in the following schema, where $\uparrow$
($\downarrow$) is an isotone (antitone) position:
$$(\uparrow\otimes\uparrow), (\uparrow\slash\downarrow), (\downarrow\bs\uparrow), 
(\uparrow\oplus\uparrow),(\uparrow\oslash\downarrow),(\downarrow\obslash\uparrow)$$
In other words, the following inference rules are valid.
\begin{equation}
\label{monoproduct}
\begin{array}{c@{\qquad}c}
\infer[]{\arr{A'\otimes B'}{A\otimes B}}{\arr{A'}{A} & \arr{B'}{B}} & 
\infer[]{\arr{A\oplus B}{A'\oplus B'}}{\arr{A}{A'} & \arr{B}{B'}}
\end{array}
\end{equation}
\begin{equation}
\label{monoimpl}
\begin{array}{c}
\infer[]{\arr{A\bs B}{A'\bs B'}}{\arr{A'}{A} & \arr{B}{B'}} \ 
\infer[]{\arr{A'\oslash B'}{A\oslash B}}{\arr{A'}{A} & \arr{B}{B'}} \ 
\infer[]{\arr{B/A}{B'/A'}}{\arr{A'}{A} & \arr{B}{B'}}  \ 
\infer[]{\arr{B'\obslash A'}{B\obslash A}}{\arr{A'}{A} & \arr{B}{B'}}
\end{array}
\end{equation}
\paragraph*{Interaction: distributivity principles} As we saw above,
one could extend the inferential capabilities of this minimal system
by adding postulates of associativity and/or commutativity
for $\otimes$ and $\oplus$. From a substructural perspective, each of
these options destroys structure-sensitivity for a particular dimension
of grammatical organization: word order in the case of commutativity,
constituent structure in the case of associativity. In \LG\, there is
an alternative which leaves the sensitivity for linear order and
phrasal structure intact: instead of considering structural options
for the individual $\otimes$ and $\oplus$ families, one can
consider \emph{interaction} principles for the communication between them.
We will consider the following group.
\begin{equation}
\label{grishindistrib}
\begin{array}{l@{\qquad}l}
(A\obslash B)\otimes C\Ra A\obslash(B\otimes C) &
C\otimes(B\oslash A)\Ra (C\otimes B)\oslash A\\
C\otimes(A\obslash B)\Ra A\obslash(C\otimes B) &
(B\oslash A)\otimes C\Ra (B\otimes C)\oslash A\\
\end{array}
\end{equation}
These postulates have come to be called \emph{linear distributivity principles}
(e.g.~\cite{Cockett96prooftheory}): linear, because they respect resources (no material gets
copied). \citeN{moot07display} models the adjunction operation of Tree Adjoining Grammars
using the interaction principles of (\ref{grishindistrib}) and shows how through this
modeling the mildly
context-sensitive patterns of (\ref{mcs}) can be obtained within \LG.

\subsection{Arrows: LG as a deductive system}\label{lgarrow}
In his \cite{lambek88}, Lambek studies the Syntactic Calculus 
from a categorical perspective.
Types are seen as the objects of a category and one studies morphisms
between these objects, arrows $\arrow{f}{A}{B}$. For each $A$, there is an
identity arrow $1_A$. Then there are inference rules to produce
new arrows from arrows already obtained. Among these is the composition $g\circ f$,
defined when $\mathit{dom}(g)=\mathit{cod}(f)$. Composition is associative, i.e.~one has the
equation $f\circ(g\circ h)=(f\circ g)\circ h$. Also,
$f\circ 1_A = f = 1_B\circ f$, where $\arrow{f}{A}{B}$.
\begin{equation}
\label{preorderarrows}
\arrow{1_A}{A}{A} \qquad \infer[]{\arrow{g\circ f}{A}{C}}{\arrow{f}{A}{B} & \arrow{g}{B}{C}}
\end{equation}
In this paper, we will not pursue the categorical interpretation of \LG:
our emphasis in the following sections is on the \emph{sequent} calculus for this logic,
the term language coding sequent proofs, and the correspondence between these proofs
and proof nets. Our aim in this section is simply to have a handy language for naming
proofs in the deductive presentation, and to use this in \S\ref{simpleseq} to establish the
equivalence between the deductive and the sequent presentations.

To obtain \textbf{aLG}, one adds to (\ref{preorderarrows}) further rules of inference for the residuation
principles and their duals. (Omitting type subscripts $\overa_{A,B,C}f$ for legibility\ldots)
\begin{equation}
\label{dualresverta}
\begin{array}{c@{\qquad}c}
\infer[]{\arrow{\overa{f}}{A}{C/B}}{\arrow{f}{A\otimes B}{C}} &
\infer[]{\arrow{\undera{f}}{B}{A\bs C}}{\arrow{f}{A\otimes B}{C}}\\
\end{array}
\end{equation}
\begin{equation}
\label{dualresvertb}
\begin{array}{c@{\qquad}c}
\infer[]{\arrow{\overai{g}}{A\otimes B}{C}}{\arrow{g}{A}{C/B}} &
\infer[]{\arrow{\underai{g}}{A\otimes B}{C}}{\arrow{g}{B}{A\bs C}}\\
\end{array}
\end{equation}
\begin{equation}
\label{dualresvertc}
\begin{array}{c@{\qquad}c}
\infer[]{\arrow{\obslasha{f}}{B\obslash C}{A}}{\arrow{f}{C}{B\oplus A}} &
\infer[]{\arrow{\oslasha{f}}{C\oslash A}{B}}{\arrow{f}{C}{B\oplus A}}\\
\end{array}
\end{equation}
\begin{equation}
\label{dualresvertd}
\begin{array}{c@{\qquad}c}
\infer[]{\arrow{\obslashai{g}}{C}{B\oplus A}}{\arrow{g}{B\obslash C}{A}} &
\infer[]{\arrow{\oslashai{g}}{C}{B\oplus A}}{\arrow{g}{C\oslash A}{B}}\\
\end{array}
\end{equation}
As remarked above, the Lambek-Grishin calculus exhibits two involutive symmetries, at the
level of types and proofs: a left-right symmetry $\cdot^{\natural}$ and an arrow
reversing symmetry $\cdot^{\dag}$ such that
\begin{equation}\label{twosymmetries}
\arrow{f^\natural}{A^\natural}{B^\natural}\quad\textrm{iff}\quad\arrow{f}{A}{B}\quad\textrm{iff}
\quad\arrow{f^\dag}{B^\dag}{A^\dag}
\end{equation}
with, on the type level, the translation tables below (abbreviating a long list of defining
equations $(A\otimes B)^\natural\stackrel{.}{=}B^\natural\otimes A^\natural,
(B\otimes A)^\natural\stackrel{.}{=}A^\natural\otimes B^\natural$, \ldots)
\[\infer=[\natural]{B\otimes A\quad B\bs A \quad B\oplus A \quad B\obslash A}{A\otimes B\quad A/B \quad A\oplus B\quad A\oslash B}
\qquad
\infer=[\dag]{B\obslash A\quad B\oplus A\quad A\oslash B}{A/B\quad A\otimes B\quad B\bs A}\]
and on the level of proofs $(1_A)^\natural = 1_{A^\natural}$, $(g\circ f)^{\natural} = g^\natural \circ f^\natural$,
$(1_A)^\dag = 1_{A^\dag}$, $(g \circ f)^\dag = f^\dag \circ g^\dag$, and the list of defining equations
$(\undera f)^\natural\stackrel{.}{=} \overa f^\natural$, $(\undera f)^\dag \stackrel{.}{=}\oslasha f^\dag$, \ldots\ 
corresponding to the translation tables above.

The distributivity principles, in \textbf{aLG}, take the form of
extra axioms (primitive arrows). Below arrows \textbf{d}, \textbf{b} for
the interaction between $\obslash$ and $\otimes$.
For the left-right symmetric pair $\textbf{d}^{\natural}$, $\textbf{q}^{\natural}$ we write $\textbf{b}$, $\textbf{p}$
\begin{equation}
\label{grishin}
\begin{array}{l}
\arrow{\textbf{d}_{A,B,C}}{(A\obslash B)\otimes C}{A\obslash(B\otimes C)}\\
\arrow{\textbf{q}_{A,B,C}}{C\otimes(A\obslash B)}{A\obslash(C\otimes B)}
\end{array}
\end{equation}
To establish the equivalence between \textbf{aLG} and the sequent
calculus \textbf{sLG}, to be discussed in the next section, we will use the fact that the
monotonicity rules are derived rules of inference of \textbf{aLG}. For example, $f/g$ can
be defined as in (\ref{monoslash}) below.
\begin{equation}
\label{monoslash}
\begin{array}{c}
\infer[]{\arrow{f/g}{A/B'}{A'/B}}{\arrow{f}{A}{A'} & \arrow{g}{B}{B'}}\\[2ex]
f/g \stackrel{.}{=} (\overa(f\circ(\overai 1_{A/B})))) \circ (\overa\underai((\undera\overai 1_{A/B'})\circ g))
\end{array}
\end{equation}
Similarly, for the distributivity postulates, we will rely on a rule form,
which for \textbf{d} would be
\begin{equation}
\label{grishrule}
\infer[]{\arr{A\obslash B}{D/C}}{
\arr{B\otimes C}{A\oplus D}}
\end{equation}
The inference rule (\ref{grishrule}) is derived as shown in (\ref{grishinruleproof}).
\begin{equation}
\label{grishinruleproof}
\infer[]{\arrow{\overa((\obslasha f)\circ\textbf{d}_{A,B,C})}{A\obslash B}{D/C}}{
\infer[]{\arrow{(\obslasha f)\circ\textbf{d}_{A,B,C}}{(A\obslash B)\otimes C}{D}}{
\arrow{\textbf{d}_{A,B,C}}{(A\obslash B)\otimes C}{A\obslash(B\otimes C)} &
\infer[]{\arrow{\obslasha f}{A\obslash(B\otimes C)}{D}}{\arrow{f}{B\otimes C}{A\oplus D}}}}
\end{equation}

\section{Display sequent calculus and proof nets}\label{seqnets}
Is there a decision procedure to determine whether $A\Ra B$ holds?
In the presence of \emph{expanding} patterns as we saw them in (\ref{coapplication}),
this is not immediately clear. For the language with $/,\otimes,\bs$, the key result of Lambek's
original papers was to establish decidability by applying Gentzen's method: the
Syntactic Calculus is recast as a sequent calculus; for the sequent
presentation one then shows that the Cut rule (the sequent form of transitivity)
is admissible; backward-chaining, cut-free proof search then yields the
desired decision procedure. 

In \S\ref{simpleseq} below, we work through a similar agenda
for \LG. We introduce \textbf{sLG}, a sequent system for the Lambek-Grishin calculus
in the style of Display Logic \cite{gore}, and show that it is equivalent to
\textbf{aLG}. The sequent presentation enjoys Cut Elimination; decidability follows.
Sequent proof search, though decidable, remains suboptimal in that it allows 
a great many derivations for what in effect one would like to consider as `the same' proof.
In \S\ref{nets}, we introduce proof nets for \LG, and show how
these nets remove the spurious forms of non-determinism of sequent
proof search.

\subsection{sLG: display sequent calculus}\label{displayseq}\label{simpleseq}
The arrows of \textbf{aLG} are morphisms between \emph{types}. In the sequent calculus,
derivability is a relation between \emph{structures} built from types. We will
present the sequent calculus for \LG\ in the format of a Display Logic
(see \cite{gore} for a comprehensive display logical view on the substructural
landscape). The characteristic feature of Display Logic is that for
every logical connective, there is a corresponding structural connective.
We use the same symbols for the logical operations and their structural counterparts;
structural operations are marked off by centerdots. Below the grammar for
input (sequent left hand side), and output structures (sequent rhs). 

\[\begin{array}{l@{\quad::=\quad}l}
\mathcal{I} & \mathcal{F}\mid
	\mathcal{I}\cdot\otimes\cdot\mathcal{I}
	\mid
	\mathcal{I}\cdot\oslash\cdot\mathcal{O}
	\mid
	\mathcal{O}\cdot\obslash\cdot\mathcal{I}\\
\mathcal{O} & \mathcal{F} \mid
	\mathcal{O}\cdot\oplus\cdot\mathcal{O}
	\mid
	\mathcal{I}\cdot\bs\cdot\mathcal{O}
	\mid
	\mathcal{O}\cdot\slash\cdot\mathcal{I}\\
\end{array}
\]
The rules of \textbf{sLG} come in three groups: the identity group (Axiom, Cut), the
structural group (Display Postulates, Distributivity Postulates), and the logical
group (left and right introduction rules for the logical connectives).
Variables $X,Y,Z$ in these rules range over structures, input or output, depending
on whether they appear left or right of the sequent arrow.
\paragraph*{Axiom, Cut}
\begin{equation}
\label{axcut}
\infer[\textrm{Ax}]{\seq{A}{A}}{} \qquad \infer[\textrm{Cut}]{\seq{X}{Y}}{\seq{X}{A} & \seq{A}{Y}}
\end{equation}
\paragraph*{Display postulates} The (dual) residuation principles are formulated at
the structural level. These rules ensure that any formula constituent of a sequent
can be displayed as the single occupant of the sequent lhs or rhs---hence the name.
\newcommand{\strutt}{\rule[-.7ex]{0cm}{3ex}}
\begin{equation}
\label{dualresvert}
\begin{array}{c@{\qquad}c}
\infer=[rp]{\strutt\seq{Y}{X\bsS Z}}{\infer=[rp]{\strutt\seq{X\otimesS Y}{Z}}{\strutt\seq{X}{Z\slashS Y}}} &
\infer=[drp]{\strutt\seq{Z\oslashS X}{Y}}{\infer=[drp]{\strutt\seq{Z}{Y\oplusS X}}{\strutt\seq{Y\obslashS Z}{X}}}
\end{array}
\end{equation}
\paragraph*{Distributivity postulates} The linear distributivities motivate the
choice for a \emph{display} sequent calculus. The distributivity postulates, in
their rule form of (\ref{grishrule}), in the sequent format
become \emph{structural} rules. In a Gentzen-style sequent calculus, formulating such
structural rules would be impossible: one only has structural punctuation marks for
$\otimes$ and $\oplus$ (the antecedent and succedent comma). But one could not formulate 
(\ref{grishrule}) as a \emph{logical} rule either: it introduces two operations
simultaneously.
\begin{equation}
\label{grishinruleseq}
\begin{array}[t]{c@{\qquad\qquad}c}
\infer[G1]{\nd{Z\obslashS X}{W\slashS Y}}{\nd{X\otimesS Y}{Z\oplusS W}}
&
\infer[G3]{\nd{Y\oslashS W}{X\bsS Z}}{\nd{X\otimesS Y}{Z\oplusS W}}\\[2ex]
\infer[G2]{\nd{Z\obslashS Y}{X\bsS W}}{\nd{X\otimesS Y}{Z\oplusS W}}
&
\infer[G4]{\nd{X\oslashS W}{Z\slashS Y}}{\nd{X\otimesS Y}{Z\oplusS W}}\\
\end{array}
\end{equation}
\paragraph*{Logical rules} For each connective there is a left and a right introduction rule.
One of these is a one-premise rewrite rule, exchanging the logical connective for its
structural counterpart; the other rule puts together a complex formula alongside
the matching complex structure.
\paragraph*{Rewrite rules} $\$\in\{\otimes,\oslash,\obslash\}$, $\#\in\{\oplus,\bs,\slash\}$.

\begin{equation}\label{rewriterules}
\infer[\$ L]{\seq{A\mathbin{\$}B}{Y}}{\seq{A\Struct{\$}B}{Y}}
\qquad
\infer[\# R]{\seq{X}{A\mathbin{\#}B}}{\seq{X}{A\Struct{\#}B}}
\end{equation}
The rewrite rules are invertible. As an example, compare $(\otimes L)$ and $(\otimes L)^{-1}$.
\begin{equation}
\label{invertibleotimes}
\infer[\otimes L]{\seq{A\otimes B}{Y}}{\seq{A\otimesS B}{Y}}
\qquad
\infer[\textrm{Cut}]{\seq{A\otimesS B}{Y}}{\infer[\otimes R]{\strutt\seq{A\otimesS B}{A\otimes B}}{\strutt\seq{A}{A} & \strutt\seq{B}{B}} & \seq{A\otimes B}{Y}}
\end{equation}
\paragraph*{Two premise rules} The $(\slash L)$, $(\obslash R)$ rules are left-right symmetric.
\begin{equation}
\label{monoproductb}
\begin{array}{c@{\qquad}c}
\infer[\otimes R]{\seq{X\otimesS Y}{A\otimes B}}{\seq{X}{A} & \seq{Y}{B}} & 
\infer[\oplus L]{\seq{A\oplus B}{X\oplusS Y}}{\seq{A}{X} & \seq{B}{Y}}
\end{array}
\end{equation}
\begin{equation}
\label{monoimplb}
\begin{array}{c@{\qquad}c}
\infer[\bs L]{\seq{A\bs B}{X\bsS Y}}{\seq{X}{A} & \seq{B}{Y}} & 
\infer[\oslash R]{\seq{X\oslashS Y}{A\oslash B}}{\seq{X}{A} & \seq{B}{Y}}
\end{array}
\end{equation}

\newcommand{\stf}[1]{#1^{\circ}}
\subsubsection*{Equivalence} For every arrow $\arrow{f}{A}{B}$, there is a sequent proof $\seq{A}{B}$. For every sequent
proof $\seq{X}{Y}$, there is an arrow $\arrow{f}{\stf{X}}{\stf{Y}}$, where ${\stf{X}},{\stf{Y}}$ are the formulas
obtained from $X,Y$ by replacing the structural connectives by their logical counterparts.

\paragraph*{From arrows to sequent proofs} $1_A$ and composition $g\circ f$ are immediate.
We use the invertibility of the rewrite rules to prove the residuation/adjoints laws in the
sequent calculus. Below, as an example, a sequent proof for $\overa{f}$.
\begin{equation}
\label{arrowtoseq}
\infer[]{\arrow{\overa{f}}{A}{C/B}}{\arrow{f}{A\otimes B}{C}}
\quad\leadsto\quad
\infer[\slash R]{\seq{A}{C/B}}{\infer[\textit{rp}]{\seq{A}{C\slashS B}}
{\infer[(\otimes L)^{-1}]{\seq{A\otimesS B}{C\textcolor{white}{\slashS}}}{\seq{A\otimes B}{C\textcolor{white}{\slashS}}}}}
\end{equation}

\paragraph*{From sequent proofs to arrows} Under the mapping $\stf{\cdot}$ Cut turns into composition of arrows,
the (dual) display postulates into the (dual) residuation rules, and the distributivity
postulates into the rule form of the arrows $\textbf{d, q, b, p}$, which in (\ref{grishinruleproof})
we have shown to be derivable
in \textbf{aLG}. For the logical group, the premise and conclusion
of the rewrite rules are identified. The two-premise logical rules become the monotonicity
rules --- derivable rules of inference in \textbf{aLG} as we saw.
\paragraph*{Cut Elimination, decidability} \cite{moortgat07sym} In \textbf{sLG}, Cut is an admissible rule: every theorem
has a cut-free derivation.

Decidability is a nice property to have. Yet, the astute reader at this point may feel
disappointed: the goal-driven, backward-chaining, cut-free proof search of the decision procedure
presupposes that the \emph{structure} of the goal sequent is given. Parsing, as it is
standardly understood, means deciding whether a string is well-formed, and assigning it
a proper structure. Here, to start backward-chaining sequent proof search, we have to
assume that the correct structure is already given. A generate-and-test
approach, obviously, is not feasible here: the number of binary bracketings over
a string of length $n$ being the Catalan number $C_n$. We haven't addressed the
parsing problem, in other words. Turning to \emph{proof nets} in \S\ref{nets}, this
situation will change: the construction algorithm for \LG\ nets will work in a
\emph{data-driven} mode, effectively \emph{computing} the structure of the
goal sequent. 

\subsection{Proof nets}\label{nets}
Proof nets are a graphical way of representing proofs, introduced
first for linear logic \cite{Girard}.  Proof nets can either be seen as a sort of ``parallellized'' sequent proofs or as a sort of multi-conclusion natural deduction.
Proof nets are defined as a subclass of a larger class of graphs called proof structures. Where proof nets correspond to sequent proofs, proof structures in general may not, but we can distinguish proof nets from other proof structures based only on properties of the graph. 

The proof nets for the Lambek-Grishin calculus we present in this
section are a simple extension of the proof nets for the multimodal
Lambek calculus of \cite{mp}.
A proof structure is a (hyper)graph where the vertices are labeled by
formulas and the edges connect these formulas. In what follows we will
often speak of formula occurrences (or simply \emph{formulas} if there is no
possibility of confusion) instead of vertices labeled by formulas. The
hyperedges correspond to the logical rules, linking the active
formulas and the main formula of the rule and keeping track of whether
one is dealing with a non-invertible two-premise rule
or with an invertible one-premise rule. We'll call these \emph{tensor}
and \emph{cotensor} links respectively.

\subsubsection*{Proof structures and abstract proof structures}

\begin{definition} A link is a tuple $\langle t,p,c,m \rangle$ where

\begin{itemize}
\item $t$ is the type of the link --- tensor or cotensor
\item $p$ is the list of premisses of the link,
\item $c$ is the list of conclusions of the link,
\item $m$, the main vertex/formula of the link, is either a member of $p$, a member of $c$ or the constant ``nil''.
\end{itemize}

In case $m$ is a member of $p$ we speak of a \emph{left} link
(corresponding to the left rules of the sequent calculus, where the
main formula of the link occurs in the antecedent) and in case $m$ is
a member of $c$ we speak of a \emph{right} link.
\end{definition}

Graphically, links are displayed as shown below. A central node links
together the premisses and conclusions of the link; when we need to
refer to the connections between the central node and the vertices, we
will call them its \emph{tentacles}. The interior of
this central node is white for a tensor link and black for a
cotensor link. The premisses are drawn, in left-to-right order, above
the central node and the conclusions, also in left-to-right order, are
drawn below it. The main formula of cotensor links is drawn as an
arrow to the member of the premisses or the conclusions which is the
main formula of the link. The main formula of tensor links are not
distinguished visually, but can be determined by inspection of the
formula labels.
  
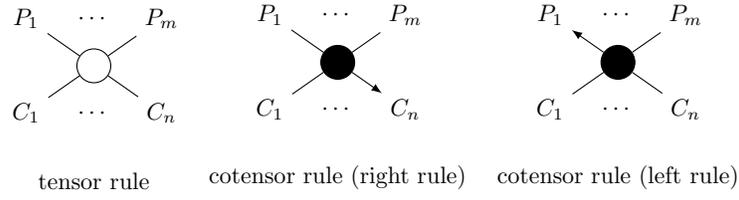
\begin{center}
\begin{tabular}{cccccc}
\begin{tikzpicture}[node distance=3mm]
\node (h1) at (0,1.4) {$P_1$};
\node (h2) at (1,1.4) {$\cdots$};
\node (h3) at (2,1.4) {$P_m$};
\node[tns] (cent1) at (1,0.7) {};
\node (c1) at (0,0) {$C_1$};
\node (c2) at (1,0) {$\cdots$};
\node (c3) at (2,0) {$C_n$};
\draw (c1) -- (cent1);
\draw (c3) -- (cent1);
\draw (h1) -- (cent1);
\draw (h3) -- (cent1);
\node (txt) at (1,-1) {tensor rule};
\end{tikzpicture} & &
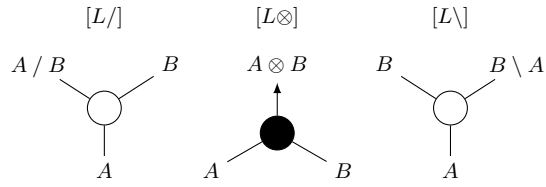
\begin{tikzpicture}[node distance=3mm]
\node (h1) at (0,1.4) {$P_1$};
\node (h2) at (1,1.4) {$\cdots$};
\node (h3) at (2,1.4) {$P_m$};
\node[par] (cent1) at (1,0.7) {};
\node (c1) at (0,0) {$C_1$};
\node (c2) at (1,0) {$\cdots$};
\node (c3) at (2,0) {$C_n$};
\draw (c1) -- (cent1);
\path[>=latex,->] (cent1) edge (c3);
\draw (h1) -- (cent1);
\draw (h3) -- (cent1);
\node (txt) at (1,-1) {cotensor rule (right rule)};
\end{tikzpicture} & &
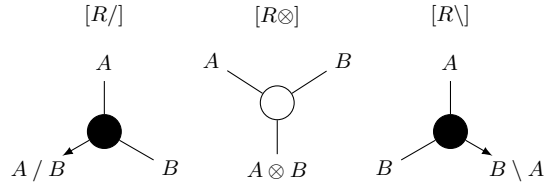
\begin{tikzpicture}[node distance=3mm]
\node (h1) at (0,1.4) {$P_1$};
\node (h2) at (1,1.4) {$\cdots$};
\node (h3) at (2,1.4) {$P_m$};
\node[par] (cent1) at (1,0.7) {};
\node (c1) at (0,0) {$C_1$};
\node (c2) at (1,0) {$\cdots$};
\node (c3) at (2,0) {$C_n$};
\draw (c1) -- (cent1);
\draw (cent1) -- (c3);
\path[>=latex,->] (cent1) edge (h1);
\draw (h3) -- (cent1);
\node (txt) at (1,-1) {cotensor rule (left rule)};
\end{tikzpicture}
\end{tabular}
\end{center}

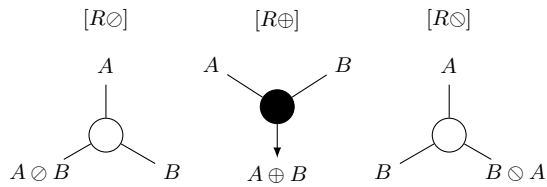
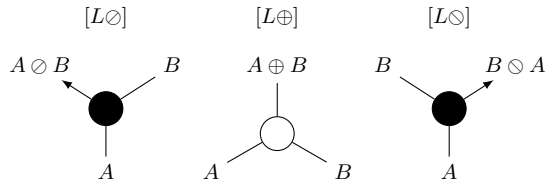
\begin{figure}
\begin{center}
\begin{tabular}{ccc}
\multicolumn{3}{c}{\textbf{Lambek connectives --- hypothesis}} \\[3mm]
\begin{tikzpicture}
\node (lab) at (3em,7em) {$[L \ldr]$};
\node (ab) at (3em,0.0em) {$A$};
\node (a) at (0,4.8em) {$\smash{A\ldr B}\rule{0pt}{1.3ex}$};
\node (b) at (6em,4.8em) {$B_{\rule{0pt}{1.2ex}}$};
\node[tns] (c) at (3em,2.868em) {};
\draw (c) -- (ab);
\draw (c) -- (a);
\draw (c) -- (b);
\end{tikzpicture} &
\begin{tikzpicture}
\node (lab) at (3em,7em) {$[L \lpr]$};
\node (ab) at (3em,4.8em) {$A\lpr B_{\rule{0pt}{1.2ex}}$};
\node (a) at (0,0) {$A$};
\node (b) at (6em,0) {$B$};
\node[par] (c) at (3em,1.732em) {};
\path[>=latex,->] (c) edge (ab);
\draw (c) -- (a);
\draw (c) -- (b);
\end{tikzpicture} &
\begin{tikzpicture}
\node (lab) at (3em,7em) {$[L \ldl]$};
\node (ab) at (3em,0.0em) {$A$};
\node (a) at (0,4.8em) {$B_{\rule{0pt}{1.2ex}}$};
\node (b) at (6em,4.8em) {$\smash{B\ldl A}\rule{0pt}{1.3ex}$};
\node[tns] (c) at (3em,2.868em) {};
\draw (c) -- (ab);
\draw (c) -- (a);
\draw (c) -- (b);
\end{tikzpicture} \\[3mm]
\multicolumn{3}{c}{\textbf{Lambek connectives --- conclusion}} \\[3mm]
\begin{tikzpicture}
\node (lab) at (3em,7em) {$[R \ldr]$};
\node (ab) at (3em,4.8em) {$A_{\rule{0pt}{1.2ex}}$};
\node (a) at (0,0) {$\smash{A\ldr B}\rule{0pt}{1.3ex}$};
\node (b) at (6em,0) {$B_{\rule{0pt}{1.2ex}}$};
\node[par] (c) at (3em,1.732em) {};
\draw (c) -- (ab);
\path[>=latex,->] (c) edge (a);
\draw (c) -- (b);
\end{tikzpicture} &
\begin{tikzpicture}
\node (lab) at (3em,7em) {$[R \lpr]$};
\node (ab) at (3em,0.0em) {$A\lpr B$};
\node (a) at (0,5.0em) {$A$};
\node (b) at (6em,5.0em) {$B$};
\node[tns] (c) at (3em,3.068em) {};
\draw (c) -- (ab);
\draw (c) -- (a);
\draw (c) -- (b);
\end{tikzpicture} &
\begin{tikzpicture}
\node (lab) at (3em,7em) {$[R \ldl]$};
\node (ab) at (3em,4.8em) {$A_{\rule{0pt}{1.2ex}}$};
\node (a) at (0,0) {$B_{\rule{0pt}{1.2ex}}$};
\node (b) at (6em,0) {$\smash{B\ldl A}\rule{0pt}{1.3ex}$};
\node[par] (c) at (3em,1.732em) {};
\draw (c) -- (ab);
\draw (c) -- (a);
\path[>=latex,->] (c) edge (b);
\end{tikzpicture} \\[3mm]
\multicolumn{3}{c}{\textbf{Grishin connectives --- hypothesis}} \\[3mm]
\begin{tikzpicture}
\node (lab) at (3em,7em) {$[L \gdr]$};
\node (ab) at (3em,0.0em) {$A$};
\node (a) at (0,4.8em) {$\smash{A\gdr B}\rule{0pt}{1.3ex}$};
\node (b) at (6em,4.8em) {$B_{\rule{0pt}{1.2ex}}$};
\node[par] (c) at (3em,2.868em) {};
\draw (c) -- (ab);
\path[>=latex,->] (c) edge (a);
\draw (c) -- (b);
\end{tikzpicture} &
\begin{tikzpicture}
\node (lab) at (3em,7em) {$[L \gpr]$};
\node (ab) at (3em,4.8em) {$A\gpr B_{\rule{0pt}{1.2ex}}$};
\node (a) at (0,0) {$A$};
\node (b) at (6em,0) {$B$};
\node[tns] (c) at (3em,1.732em) {};
\draw (c) -- (ab);
\draw (c) -- (a);
\draw (c) -- (b);
\end{tikzpicture} &
\begin{tikzpicture}
\node (lab) at (3em,7em) {$[L \gdl]$};
\node (ab) at (3em,0.0em) {$A$};
\node (a) at (0,4.8em) {$B_{\rule{0pt}{1.2ex}}$};
\node (b) at (6em,4.8em) {$\smash{B\gdl A}\rule{0pt}{1.3ex}$};
\node[par] (c) at (3em,2.868em) {};
\draw (c) -- (ab);
\draw (c) -- (a);
\path[>=latex,->] (c) edge (b);
\end{tikzpicture} \\[3mm]
\multicolumn{3}{c}{\textbf{Grishin connectives --- conclusion}} \\[3mm]
\begin{tikzpicture}
\node (lab) at (3em,7em) {$[R \gdr]$};
\node (ab) at (3em,4.8em) {$A_{\rule{0pt}{1.2ex}}$};
\node (a) at (0,0) {$\smash{A\gdr B}\rule{0pt}{1.3ex}$};
\node (b) at (6em,0) {$B_{\rule{0pt}{1.2ex}}$};
\node[tns] (c) at (3em,1.732em) {};
\draw (c) -- (ab);
\draw (c) -- (a);
\draw (c) -- (b);
\end{tikzpicture} &
\begin{tikzpicture}
\node (lab) at (3em,7em) {$[R \gpr]$};
\node (ab) at (3em,0.0em) {$A\gpr B$};
\node (a) at (0,5.0em) {$A$};
\node (b) at (6em,5.0em) {$B$};
\node[par] (c) at (3em,3.068em) {};
\path[>=latex,->]  (c) edge (ab);
\draw (c) -- (a);
\draw (c) -- (b);
\end{tikzpicture} &
\begin{tikzpicture}
\node (lab) at (3em,7em) {$[R \gdl]$};
\node (ab) at (3em,4.8em) {$A_{\rule{0pt}{1.2ex}}$};
\node (a) at (0,0) {$B_{\rule{0pt}{1.2ex}}$};
\node (b) at (6em,0) {$\smash{B\gdl A}\rule{0pt}{1.3ex}$};
\node[tns] (c) at (3em,1.732em) {};
\draw (c) -- (ab);
\draw (c) -- (a);
\draw (c) -- (b);
\end{tikzpicture} \\
\end{tabular}
\end{center}
\caption{Links for proof structures of the Lambek-Grishin calculus}
\label{fig:links}
\end{figure}

Figure~\ref{fig:links} shows the links for the Lambek-Grishin
calculus: there are two links for each connective, one link where the
main formula is a premiss of the link (a left link) and one link where
the main formula is a conclusion of the link (a right link). The
symmetry between the Lambek connectives and the Grishin connectives is
immediately clear: the links for the Grishin connectives are up-down
symmetric versions of the links for the Lambek connectives.

\begin{definition} A \emph{proof structure} $\langle S, \mathcal{L}\rangle$ is a finite set of formula occurrences $S$ and a set of links $\mathcal{L}$ from those shown in Figure~\ref{fig:links} such that.

\begin{itemize}
\item each formula is at most once the premiss of a link,
\item each formula is at most once the conclusion of a link.
\end{itemize}

Formulas which are not the conclusion of any link are called the \emph{hypotheses} of the proof structure. Formulas which are not the premiss of any link are called the \emph{conclusions} of the proof structure. 

We will say that a proof structure with hypotheses $H_1,\ldots,H_m$ and conclusions $C_1,\ldots,C_n$ is a proof structure of $\seq{H_1,\ldots,H_m}{C_1,\ldots C_n}$.
\end{definition}


\begin{example} Figure~\ref{fig:ex_unfold} shows the hypothesis
  unfolding of $(s \gdr s) \gdl np$ and the conclusion unfolding of
  $s\ldr (np\ldl s)$. Both are obtained by simple application of the
  rules of Figure~\ref{fig:links} until we reach the atomic subformulas.

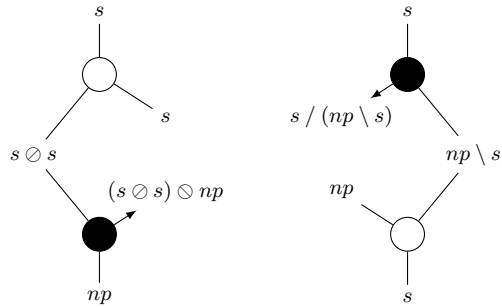
\begin{figure}
\begin{center}
\begin{tikzpicture}
\node (lab) at (23em,13.0em) {$s$};
\node (la) at (26em,6.5em) {$np\ldl s$};
\node (lb) at (20em,8.2em) {$s\ldr (np\ldl s)$};
\node[par] (lc) at (23em,10.132em) {};
\draw (lc) -- (lab);
\draw (lc) -- (la);
\path[>=latex,->] (lc) edge (lb);
\node (lla) at (20em,4.8em) {$np$};
\node[tns] (llc) at (23em,2.868em) {};
\node (lld) at (23em,0em) {$s$};
\draw (llc) -- (lld);
\draw (llc) -- (lla);
\draw (llc) -- (la);
\node (lab) at (9em,13.0em) {$s$};
\node (la) at (12em,8.2em) {$s$};
\node (lb) at (6em,6.5em) {$s \gdr s$};
\node[tns] (lc) at (9em,10.132em) {};
\draw (lc) -- (lab);
\draw (lc) -- (la);
\draw (lc) -- (lb);
\node (llb) at (12em,4.8em) {$(s \gdr s) \gdl np$};
\node[par] (llc) at (9em,2.868em) {};
\node (lld) at (9em,0em) {$np$};
\draw (llc) -- (lld);
\draw (llc) -- (lb);
\path[>=latex,->] (llc) edge (llb);
\end{tikzpicture}
\end{center}
\caption{Lexical unfolding}
\label{fig:ex_unfold}
\end{figure}

Though the figure satisfies the conditions of being a proof structure
(note, for example, that connectedness is not a requirement, so a proof
structure is allowed to have one connected component for each of the
unfolded formulas),
it is a proof structure of $\seq{(s \gdr s) \gdl np, s, s, np}{s\ldr (np\ldl
  s), s, s, np}$. We can obtain a proof structure of $\seq{(s \gdr s) \gdl np}{s\ldr (np\ldl
  s)}$ by identifying atomic formulas (this node identification corresponds to the ``axiom
  links'' of linear logic proof nets). In this case, we choose to
  identify the top $s$ of the left subgraph with the bottom $s$ of the
  right subgraph and perform the unique choice for the remaining
  atomic formulas. The result is the proof structure
  shown in Figure~\ref{fig:ex_pn} on the left.

\begin{figure}
\begin{center}
\begin{tikzpicture}
\node (mab) at (21em,4.8em) {$s$};
\node (ma) at (24em,0.0em) {$s$};
\node (mb) at (18em,0.0em) {$s \gdr s$};
\node[tns] (mc) at (21em,1.932em) {};
\draw (mc) -- (mab);
\draw (mc) -- (ma);
\draw (mc) -- (mb);
\node (mla) at (18em,9.6em) {$np$};
\node (mlb) at (24em,9.6em) {$np \ldl s$};
\node[tns] (mlc) at (21em,7.668em) {};
\draw (mlc) -- (mab);
\draw (mlc) -- (mla);
\draw (mlc) -- (mlb);
\node (b) at (21em,14.4em) {$(s \gdr s) \gdl np$};
\node[par] (c) at (18em,12.468em) {};
\draw (c) -- (mla);
\path[>=latex,->] (c) edge (b);
\draw (c) to [out=130,in=210] (mb);
\node (gl) at (21em,-4.8em) {$s\ldr (np\ldl s)$};
\node[par] (cl) at (24em,-3.068em) {};
\draw (cl) -- (ma);
\path[>=latex,->] (cl) edge (gl);
\draw (mlb) to [out=50,in=330] (cl);
\node (mab) at (41em,4.8em) {$\apsnodei$};
\node (ma) at (44em,0.0em) {$\apsnodei$};
\node (mb) at (38em,0.0em) {$\apsnodei$};
\node[tns] (mc) at (41em,1.932em) {};
\draw (mc) -- (mab);
\draw (mc) -- (ma);
\draw (mc) -- (mb);
\node (mla) at (38em,9.6em) {$\apsnodei$};
\node (mlb) at (44em,9.6em) {$\apsnodei$};
\node[tns] (mlc) at (41em,7.668em) {};
\draw (mlc) -- (mab);
\draw (mlc) -- (mla);
\draw (mlc) -- (mlb);
\node (b) at (41em,14.4em) {$\smash{\apsnode{(s \gdr s) \gdl np}{}}\rule{0pt}{1ex}$};
\node[par] (c) at (38em,12.468em) {};
\draw (c) -- (mla);
\path[>=latex,->] (c) edge (b);
\draw (c) to [out=130,in=210] (mb);
\node (gl) at (41em,-4.8em) {$\smash{\apsnodec{s\ldr (np\ldl s)}}\rule{0pt}{1ex}$};
\node[par] (cl) at (44em,-3.068em) {};
\draw (cl) -- (ma);
\path[>=latex,->] (cl) edge (gl);
\draw (mlb) to [out=50,in=330] (cl);
\end{tikzpicture}
\end{center}
\caption{Proof structure of $\seq{(s \gdr s) \gdl np}{s\ldr (np\ldl
  s)}$ corresponding to the lexical unfolding of
  Figure~\ref{fig:ex_unfold} (left) and its corresponding abstract
  proof structure}
\label{fig:ex_pn}
\end{figure}

Let's take a closer look at this new proof structure. We have conneced the minor premiss of the implication and
co-implication links by a curve. This is due to the graphical
constraints of writing these proof nets on the plane: we want to draw the
$np\ldl s$ node \emph{below} the cotensor link at the bottom of the
figure, since it is a conclusion of this link, but would have to draw
the figure on a cylinder to make this work --- in other words,
following down a path premiss - link - conclusion does not necessarily
give a total order but can give a cyclic order on the formulas in the
proof structure; for proof \emph{nets}, these cyclic paths can only
pass through the minor premiss of a cotensor (co-)implication link. As
indicated by the drawing, the connection
to the $np\ldl s$ node from the cotensor link arrives from above,
indicating it is a conclusion of this link. Similarly, we go down from
the $s \gdr s$ node to arrive at the other cotensor link.

A comparison with the introduction rule for the implication in natural
deduction is another
way to make this clear. For the introduction rule, we hypothesise a
formula $np\ldl s$ (here, a conclusion of the cotensor rule), then
derive $s$ (here a premiss of the cotensor rule). The introduction rule then indicates
we can withdraw this hypothesis and conclude $s \ldr (np\ldl s)$, with
some indexing indicating which hypotheses are withdrawn at which
rule. In the proof structure above, the connection between the
cotensor link and the $np\ldl s$ rule plays exactly the role of this
indexing (though, since a proof structure is not necessarily a
proof, we have no guarantee yet that the introduction rule is
correctly applied; the contractions introduced later will remedy this).

With this in mind, we can verify that the proof structure in
Figure~\ref{fig:ex_pn} corresponds exactly to the one in
Figure~\ref{fig:ex_unfold} with the stated node identifications: we
have the same formula occurrences and the
links have the same premisses as well as the same conclusions.
\end{example}

So while the logical rules of a sequent proof correspond directly to
the links of a proof net, the axioms and cut rules of a sequent proof
correspond to \emph{formulas}. An axiomatic formula is a formula which
is not the main formula of any link. A cut formula is a formula which
is the main formula of two links. So on the left of Figure~\ref{fig:ex_pn}, the
$np$ formula and both $s$ formulas are axiomatic.

\begin{definition} An \emph{abstract proof structure} $\langle V, \mathcal{L},h,c\rangle$ is a set of vertices $V$, a set of (unlabeled) links $\mathcal{L}$ and two functions $h$ and $c$, such that.

\begin{itemize}
\item each formula is at most once the premiss of a link,
\item each formula is at most once the conclusion of a link,
\item $h$ is a function from the hypotheses of the abstract proof structure to formulas,
\item $c$ is a function from the conclusions of the abstract proof structure to formulas.
\end{itemize}

\end{definition}

 Note that the abstract proof structure corresponding to a two formula sequent $\seq{A}{B}$ has only a single vertex $v$, with $h(v) = A$ and $c(v) = B$.

The transformation from proof structure to abstract proof structure is
a forgetful mapping: we transform a proof structure into an abstract proof
structure by erasing all formula information on the internal vertices,
keeping only the formula labels of the hypotheses and the
conclusions. Visually, we remove the formula labels of the graph and
replace them by simple vertices ($\centerdot$) and we indicate the
results of the functions $h$ and $c$ above (resp.\ below) the vertices (those which
are hypotheses and conclusions of the abstract proof structure
respectively). As a result, we have to following four types of
vertices in an abstract proof structure.

\begin{center}
\begin{tabular}{m{5em}m{5em}m{5em}m{5em}}
\begin{center}
$\centerdot$
\end{center} & 
\begin{center}
$\apsnode{A}{}$
\end{center} & 
\begin{center}
$\apsnode{}{B}$
\end{center} &
\begin{center} 
$\apsnode{A}{B}$
\end{center} \\[4mm]
\begin{center}
internal
\end{center} & 
\begin{center}
hypothesis
\end{center} & 
\begin{center}
conclusion
\end{center} & 
\begin{center}
both
\end{center}
\end{tabular}
\end{center}

\begin{example}  Figure~\ref{fig:ex_pn} shows (on the right) the
  transformation of the proof structure on its left into an abstract
  proof structure. In the abstract proof structure, we can no longer
  distinguish which vertices are axioms: only the cotensor links still
  allow us to distinguish between the main and active vertices of the link by
  means of the arrow.
\end{example}

\begin{definition}\label{def:tree} A \emph{tree} is an acyclic, connected abstract proof structure which does not contain any cotensor links.
\end{definition}

The trees of Definition~\ref{def:tree} correspond to sequents in a rather direct way. In fact, they have the rather pleasant property of ``compiling away'' the display rules of the sequent calculus. Or, in other words, trees represent a class of sequents which is equivalent up to the display postulates. 

\begin{figure}
\begin{center}
\begin{tikzpicture}
\node (labl) at (3em,-2.5em) {$[R\ldr ]$};
\node (ab) at (3em,4.8em) {$\apsnodei$};
\node (a) at (0,9.6em) {$\smash{\apsnodeh{H}}$};
\node (b) at (6em,9.6em) {$\apsnodei$};
\node[tns] (c) at (3em,7.668em) {};
\draw (c) -- (ab);
\draw (c) -- (a);
\draw (c) -- (b);
\node (pa) at (0,0) {$\smash{\apsnodec{C}}$};
\node[par] (pc) at (3em,1.732em) {};
\draw (pc) -- (ab);
\path[>=latex,->]  (pc) edge (pa);
\draw (b) to [out=50,in=330] (pc);
\node (labm) at (13em,-2.5em) {$[L\lpr ]$};
\node (tab) at (13em,0.0em) {$\apsnodec{C}$};
\node (ta) at (10em,4.8em) {$\apsnodei$};
\node (tb) at (16em,4.8em) {$\apsnodei$};
\node[tns] (tc) at (13em,2.868em) {};
\draw (tc) -- (tab);
\draw (tc) -- (ta);
\draw (tc) -- (tb);
\node (tabx) at (13em,9.6em) {$\apsnodeh{H}$};
\node[par] (tcx) at (13em,6.532em) {};
\path[>=latex,->]  (tcx) edge (tabx);
\draw (tcx) -- (ta);
\draw (tcx) -- (tb);
\node (labr) at (23em,-2.5em) {$[R\ldl ]$};
\node (ab) at (23em,4.8em) {$\apsnodei$};
\node (a) at (26em,9.6em) {$\smash{\apsnodeh{H}}$};
\node (b) at (20em,9.6em) {$\apsnodei$};
\node[tns] (c) at (23em,7.668em) {};
\draw (c) -- (ab);
\draw (c) -- (a);
\draw (c) -- (b);
\node (pa) at (26em,0) {$\smash{\apsnodec{C}}$};
\node[par] (pc) at (23em,1.732em) {};
\draw (pc) -- (ab);
\path[>=latex,->]  (pc) edge (pa);
\draw (b) to [out=130,in=210] (pc);
\end{tikzpicture}
\end{center}
\caption{Contractions --- Lambek connectives}
\label{fig:lambek_contr}
\end{figure}
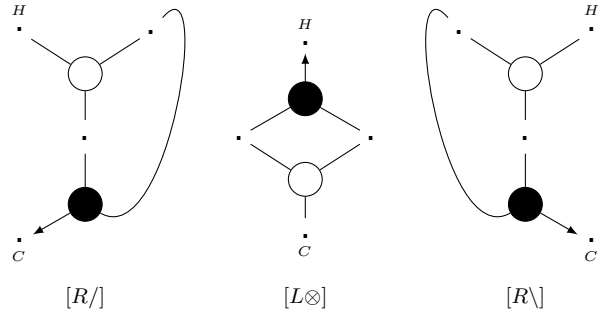

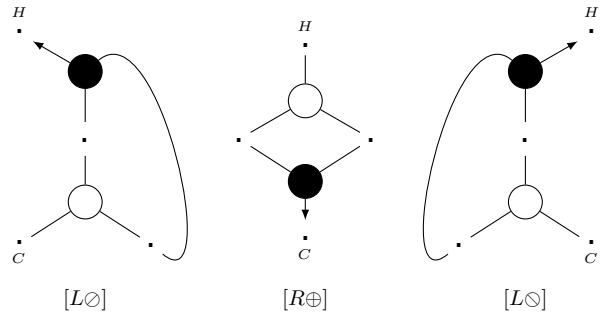
\begin{figure}
\begin{center}
\begin{tikzpicture}
\node (labl) at (3em,-2.5em) {$[L\gdr ]$};
\node (ab) at (3em,4.8em) {$\apsnodei$};
\node (a) at (0,0.0em) {$\smash{\apsnodec{C}}$};
\node (b) at (6em,0.0em) {$\apsnodei$};
\node[tns] (c) at (3em,1.932em) {};
\draw (c) -- (ab);
\draw (c) -- (a);
\draw (c) -- (b);
\node (pa) at (0,9.6em) {$\smash{\apsnodeh{H}}$};
\node[par] (pc) at (3em,7.868em) {};
\draw (pc) -- (ab);
\path[>=latex,->]  (pc) edge (pa);
\draw (b) to [out=320,in=40] (pc);
\node (labm) at (13em,-2.5em) {$[R\gpr ]$};
\node (tab) at (13em,0.0em) {$\apsnodec{C}$};
\node (ta) at (10em,4.8em) {$\apsnodei$};
\node (tb) at (16em,4.8em) {$\apsnodei$};
\node[par] (tc) at (13em,2.868em) {};
\path[>=latex,->] (tc) edge (tab);
\draw (tc) -- (ta);
\draw (tc) -- (tb);
\node (tabx) at (13em,9.6em) {$\apsnodeh{H}$};
\node[tns] (tcx) at (13em,6.532em) {};
\draw (tcx) -- (tabx);
\draw (tcx) -- (ta);
\draw (tcx) -- (tb);
\node (labr) at (23em,-2.5em) {$[L\gdl ]$};
\node (ab) at (23em,4.8em) {$\apsnodei$};
\node (a) at (26em,0.0em) {$\smash{\apsnodec{C}}$};
\node (b) at (20em,0.0em) {$\apsnodei$};
\node[tns] (c) at (23em,1.932em) {};
\draw (c) -- (ab);
\draw (c) -- (a);
\draw (c) -- (b);
\node (pa) at (26em,9.6em) {$\smash{\apsnodeh{H}}$};
\node[par] (pc) at (23em,7.868em) {};
\draw (pc) -- (ab);
\path[>=latex,->]  (pc) edge (pa);
\draw (b) to [out=220,in=140] (pc);
\end{tikzpicture}
\end{center}
\caption{Contractions --- Grishin connectives}
\label{fig:grishin_contr}
\end{figure}

\begin{definition} Given an abstract proof structure $A$, we say that $A$ \emph{contracts in one step} to $A'$, written $A\rightarrow A'$ iff $A'$ is obtained from $A$ by replacing one of the subgraphs of the form shown in Figures~\ref{fig:lambek_contr} and \ref{fig:grishin_contr} by a single vertex.

$$
\apsnode{H}{C}
$$

$H$ represents the result of the function $h$ for the indicated node (relevant only in case this node is a hypothesis of the abstract proof structure). Similarly, $C$ represent the formula assigned by the function $c$ to the indicated node. 

Given an abstract proof structure $A$ we say that $A$ \emph{contracts to} an abstract proof structure $A'$ if there is a sequence of zero or more one step contractions from $A$ to $A'$.

When we say that a \emph{proof structure} $P$ contracts to an abstract proof structure $A'$ we will mean that the underlying abstract proof structure $A$ of $P$ contracts to $A'$.
\end{definition}

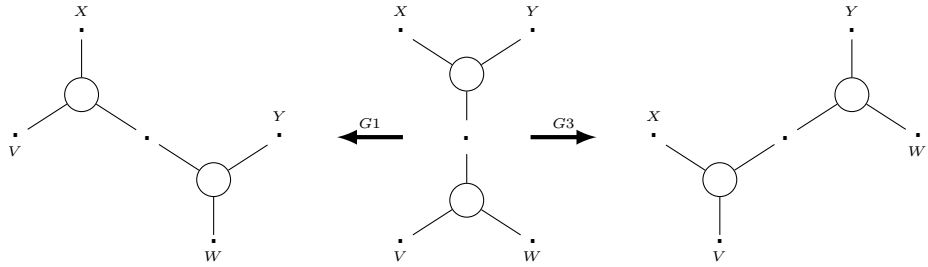
\begin{figure}
\begin{center}
\begin{tikzpicture}
\node (lab) at (3em,9.6em) {$\smash{\apsnodeh{X}}$};
\node (la) at (6em,4.8em) {$\apsnodei$};
\node (lb) at (0em,4.8em) {$\smash{\apsnodec{V}}$};
\node[tns] (lc) at (3em,6.732em) {};
\draw (lc) -- (lab);
\draw (lc) -- (la);
\draw (lc) -- (lb);
\node (llb) at (12em,4.8em) {$\smash{\apsnodeh{Y}}$};
\node[tns] (llc) at (9em,2.868em) {};
\node (lld) at (9em,0em) {$\smash{\apsnodec{W}}$};
\draw (llc) -- (lld);
\draw (llc) -- (la);
\draw (llc) -- (llb);
\node (mab) at (20.5em,4.8em) {$\apsnodei$};
\node (ma) at (23.5em,0.0em) {$\smash{\apsnodec{W}}$};
\node (mb) at (17.5em,0.0em) {$\smash{\apsnodec{V}}$};
\node[tns] (mc) at (20.5em,1.932em) {};
\draw (mc) -- (mab);
\draw (mc) -- (ma);
\draw (mc) -- (mb);
%
\node (mla) at (17.5em,9.6em) {$\smash{\apsnodeh{X}}$};
\node (mlb) at (23.5em,9.6em) {$\smash{\apsnodeh{Y}}$};
\node[tns] (mlc) at (20.5em,7.668em) {};
\draw (mlc) -- (mab);
\draw (mlc) -- (mla);
\draw (mlc) -- (mlb);
\node (lab) at (38em,9.6em) {$\smash{\apsnodeh{Y}}$};
\node (la) at (41em,4.8em) {$\smash{\apsnodec{W}}$};
\node (lb) at (35em,4.8em) {$\apsnodei$};
\node[tns] (lc) at (38em,6.732em) {};
\draw (lc) -- (lab);
\draw (lc) -- (la);
\draw (lc) -- (lb);
\node (llb) at (29em,4.8em) {$\smash{\apsnodeh{X}}$};
\node[tns] (llc) at (32em,2.868em) {};
\node (lld) at (32em,0em) {$\smash{\apsnodec{V}}$};
\draw (llc) -- (lld);
\draw (llc) -- (lb);
\draw (llc) -- (llb);
\path[ultra thick,>=latex,->] (17.6em,4.8em) edge node[above] {$\scriptstyle G1$} (14.6em,4.8em);
\path[ultra thick,>=latex,->] (23.4em,4.8em) edge node[above] {$\scriptstyle G3$} (26.4em,4.8em);
\end{tikzpicture}
\end{center}
\caption{Grishin interactions I --- ``mixed associativity''}
\label{fig:grishin_ma}
\end{figure}

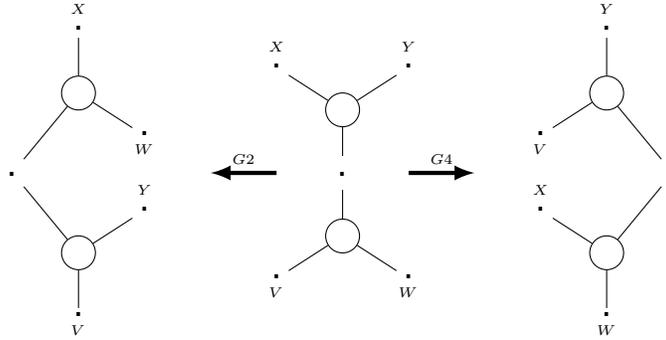
\begin{figure}[h]
\begin{center}
\begin{tikzpicture}
\node (lab) at (33em,13.0em) {$\smash{\apsnodeh{Y}}$};
\node (la) at (36em,6.5em) {$\apsnodei$};
\node (lb) at (30em,8.2em) {$\smash{\apsnodec{V}}$};
\node[tns] (lc) at (33em,10.132em) {};
\draw (lc) -- (lab);
\draw (lc) -- (la);
\draw (lc) -- (lb);
\node (lla) at (30em,4.8em) {$\smash{\apsnodeh{X}}$};
\node[tns] (llc) at (33em,2.868em) {};
\node (lld) at (33em,0em) {$\smash{\apsnodec{W}}$};
\draw (llc) -- (lld);
\draw (llc) -- (lla);
\draw (llc) -- (la);
\node (mab) at (21em,6.5em) {$\apsnodei$};
\node (ma) at (24em,1.7em) {$\smash{\apsnodec{W}}$};
\node (mb) at (18em,1.7em) {$\smash{\apsnodec{V}}$};
\node[tns] (mc) at (21em,3.632em) {};
\draw (mc) -- (mab);
\draw (mc) -- (ma);
\draw (mc) -- (mb);
%
\node (mla) at (18em,11.3em) {$\smash{\apsnodeh{X}}$};
\node (mlb) at (24em,11.3em) {$\smash{\apsnodeh{Y}}$};
\node[tns] (mlc) at (21em,9.368em) {};
\draw (mlc) -- (mab);
\draw (mlc) -- (mla);
\draw (mlc) -- (mlb);
\node (lab) at (9em,13.0em) {$\smash{\apsnodeh{X}}$};
\node (la) at (12em,8.2em) {$\smash{\apsnodec{W}}$};
\node (lb) at (6em,6.5em) {$\apsnodei$};
\node[tns] (lc) at (9em,10.132em) {};
\draw (lc) -- (lab);
\draw (lc) -- (la);
\draw (lc) -- (lb);
\node (llb) at (12em,4.8em) {$\smash{\apsnodeh{Y}}$};
\node[tns] (llc) at (9em,2.868em) {};
\node (lld) at (9em,0em) {$\smash{\apsnodec{V}}$};
\draw (llc) -- (lld);
\draw (llc) -- (lb);
\draw (llc) -- (llb);
\path[ultra thick,>=latex,->] (18em,6.5em) edge node[above] {$\scriptstyle G2$} (15em,6.5em);
\path[ultra thick,>=latex,->] (24em,6.5em) edge node[above] {$\scriptstyle G4$} (27em,6.5em);
\end{tikzpicture}
\end{center}
\caption{Grishin interactions II --- ``mixed commutativity''}
\label{fig:grishin_mc}
\end{figure}

As we saw in \S\ref{lgarrow}, to obtain expressivity beyond context-free, we are interested in \LG\ with
added interaction principles. The (rule forms of the) postulates (\S\ref{grishin})
correspond to additional rewrite rules on the abstract proof structures.
Figures~\ref{fig:grishin_ma} and \ref{fig:grishin_mc} give the rewrite rules
corresponding to the postulates \textbf{d}, \textbf{b}
and \textbf{q}, \textbf{p} respectively\footnote{These are
Grishin's Class IV interactions. His Class I can be obtained by inversing all four arrows in the two figures.}; a total of four rewrite rules ($G1$) to ($G4$). 
All four rewrite rules start from the same inital configuration and replace it
by one of the four possible configurations indicated in the figures.

\subsubsection*{Proof nets}
\begin{definition} A proof structure $P$ is a \emph{proof net} iff its underlying abstract proof structure $A$ converts to a tree using the contractions of Figures~\ref{fig:lambek_contr} and \ref{fig:grishin_contr} and the structural rules of Figures~\ref{fig:grishin_ma} and \ref{fig:grishin_mc}.
\end{definition}

\begin{example} To show that the proof structure of
 Figure~\ref{fig:ex_pn} is a proof net, we need to show it can be
 contracted to a tree. Inspection of the contractions shows that none
 of them apply, but the interaction rules do: the two tensor links
 in the center of the figure are in the right configuration for the
 interaction rules. Applying rule ($G1$)
 produces the abstract proof structure shown in
 Figure~\ref{fig:ex_pnet} on the right.

\begin{figure}
\begin{center}
\begin{tikzpicture}
\node (mab) at (11em,4.8em) {$\apsnodei$};
\node (ma) at (14em,0.0em) {$\apsnodei$};
\node (mb) at (8em,0.0em) {$\apsnodei$};
\node[tns] (mc) at (11em,1.932em) {};
\draw (mc) -- (mab);
\draw (mc) -- (ma);
\draw (mc) -- (mb);
\node (mla) at (8em,9.6em) {$\apsnodei$};
\node (mlb) at (14em,9.6em) {$\apsnodei$};
\node[tns] (mlc) at (11em,7.668em) {};
\draw (mlc) -- (mab);
\draw (mlc) -- (mla);
\draw (mlc) -- (mlb);
\node (b) at (11em,14.4em) {$\smash{\apsnode{(s \gdr s) \gdl np}{}}\rule{0pt}{1ex}$};
\node[par] (c) at (8em,12.468em) {};
\draw (c) -- (mla);
\path[>=latex,->] (c) edge (b);
\draw (c) to [out=130,in=210] (mb);
\node (gl) at (11em,-4.8em) {$\smash{\apsnodec{s\ldr (np\ldl s)}}\rule{0pt}{1ex}$};
\node[par] (cl) at (14em,-3.068em) {};
\draw (cl) -- (ma);
\path[>=latex,->] (cl) edge (gl);
\draw (mlb) to [out=50,in=330] (cl);
%
\node (mab) at (28em,9.6em) {$\apsnodei$};
\node (ma) at (31em,4.8em) {$\apsnodei$};
\node (mb) at (25em,4.8em) {$\apsnodei$};
\node[tns] (mc) at (28em,6.732em) {};
\draw (mc) -- (mab);
\draw (mc) -- (ma);
\draw (mc) -- (mb);
\node (mla) at (37em,4.8em) {$\apsnodei$};
\node (mlb) at (34em,0.0em) {$\apsnodei$};
\node[tns] (mlc) at (34em,2.868em) {};
\draw (mlc) -- (ma);
\draw (mlc) -- (mla);
\draw (mlc) -- (mlb);
\node (b) at (31em,14.4em) {$\smash{\apsnode{(s \gdr s) \gdl np}{}}\rule{0pt}{1ex}$};
\node[par] (c) at (28em,12.468em) {};
\draw (c) -- (mab);
\path[>=latex,->] (c) edge (b);
\draw (c) to [out=130,in=210] (mb);
\node (gl) at (31em,-4.8em) {$\smash{\apsnodec{s\ldr (np\ldl s)}}\rule{0pt}{1ex}$};
\node[par] (cl) at (34em,-3.068em) {};
\draw (cl) -- (mlb);
\path[>=latex,->] (cl) edge (gl);
\draw (mla) to [out=50,in=330] (cl);
\path[ultra thick,>=latex,<-] (22em,4.8em) edge node[above] {$\scriptstyle G1$} (19em,4.8em);
\end{tikzpicture}
\end{center}
\caption{Applying rule ($G1$) to the abstract proof structure of Figure~\ref{fig:ex_pn}}
\label{fig:ex_pnet}
\end{figure}
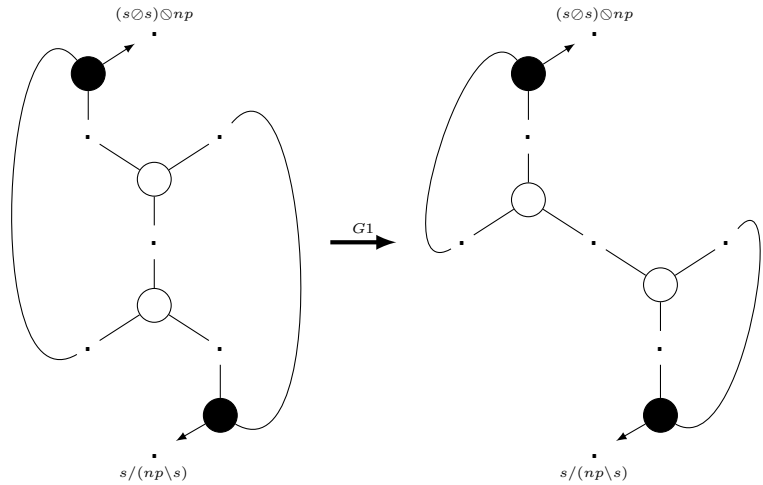

Now, we are in the right structure to contract the two cotensor links. Any order is possible.  Figure~\ref{fig:ex_pnetb} shows the
result of first applying the ($L\gdl$), then the ($R\ldr$) contraction.

\begin{figure}
\begin{center}
\begin{tikzpicture}
%
\node (mab) at (28em,9.6em) {$\apsnodei$};
\node (ma) at (31em,4.8em) {$\apsnodei$};
\node (mb) at (25em,4.8em) {$\apsnodei$};
\node[tns] (mc) at (28em,6.732em) {};
\draw (mc) -- (mab);
\draw (mc) -- (ma);
\draw (mc) -- (mb);
\node (mla) at (37em,4.8em) {$\apsnodei$};
\node (mlb) at (34em,0.0em) {$\apsnodei$};
\node[tns] (mlc) at (34em,2.868em) {};
\draw (mlc) -- (ma);
\draw (mlc) -- (mla);
\draw (mlc) -- (mlb);
\node (b) at (31em,14.4em) {$\smash{\apsnode{(s \gdr s) \gdl np}{}}\rule{0pt}{1ex}$};
\node[par] (c) at (28em,12.468em) {};
\draw (c) -- (mab);
\path[>=latex,->] (c) edge (b);
\draw (c) to [out=130,in=210] (mb);
\node (gl) at (31em,-4.8em) {$\smash{\apsnodec{s\ldr (np\ldl s)}}\rule{0pt}{1ex}$};
\node[par] (cl) at (34em,-3.068em) {};
\draw (cl) -- (mlb);
\path[>=latex,->] (cl) edge (gl);
\draw (mla) to [out=50,in=330] (cl);
\path[ultra thick,>=latex,<-] (41em,-4.8em) edge node[above] {$\scriptstyle L\gdl$} (38em,-4.8em);
%
%
\node (ma) at (44em,4.8em) {$\smash{\apsnode{(s \gdr s) \gdl np}{}}\rule{0pt}{1ex}$};
\node (mla) at (50em,4.8em) {$\apsnodei$};
\node (mlb) at (47em,0.0em) {$\apsnodei$};
\node[tns] (mlc) at (47em,2.868em) {};
\draw (mlc) -- (ma);
\draw (mlc) -- (mla);
\draw (mlc) -- (mlb);
\node (gl) at (44em,-4.8em) {$\smash{\apsnodec{s\ldr (np\ldl s)}}\rule{0pt}{1ex}$};
\node[par] (cl) at (47em,-3.068em) {};
\draw (cl) -- (mlb);
\path[>=latex,->] (cl) edge (gl);
\draw (mla) to [out=50,in=330] (cl);
\path[ultra thick,>=latex,<-] (53em,-4.8em) edge node[above] {$\scriptstyle R\ldr$} (50em,-4.8em);
%
%
\node (gl) at (56em,-4.8em) {$\smash{\apsnode{(s \gdr s) \gdl np}{s\ldr (np\ldl s)}}\rule{0pt}{1ex}$};
\end{tikzpicture}
\end{center}
\caption{Applying rule ($L\gdl$) and ($R\ldr$) contractions to the abstract proof structure of Figure~\ref{fig:ex_pnet}}
\label{fig:ex_pnetb}
\end{figure}
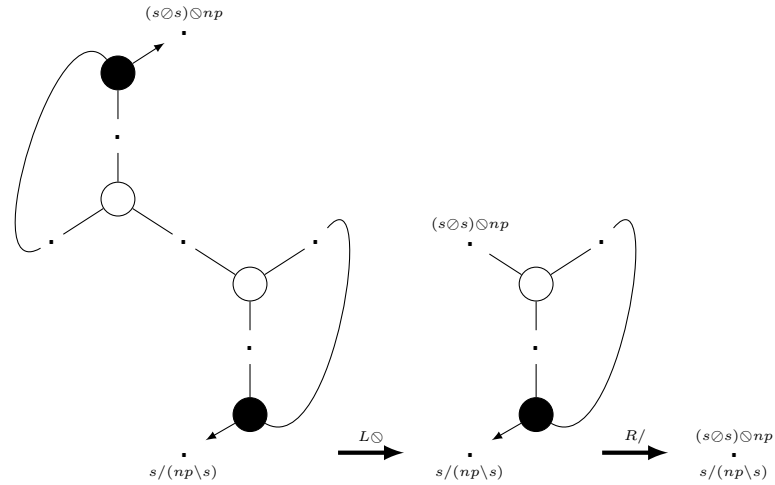

\end{example}

\begin{example}  Figure~\ref{fig:grishin_lex} shows the lexical proof
  structures for a generalized quantifier noun phrase, a transitive verb, a
  determiner and a lexical noun.

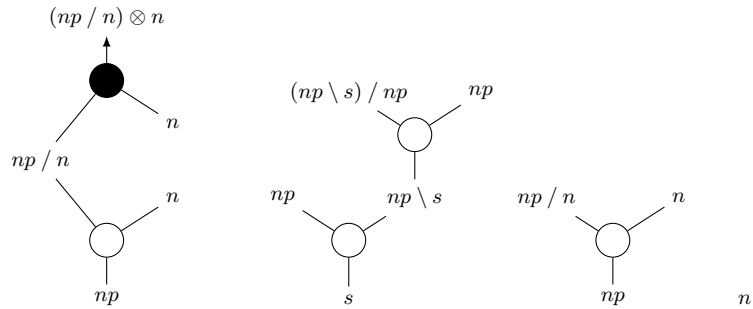
\begin{figure}[h]
\begin{center}
\begin{tikzpicture}
\node (lab) at (9em,13.0em) {$(np\ldr n)\lpr n$};
\node (la) at (12em,8.2em) {$n$};
\node (lb) at (6em,6.5em) {$np\ldr n$};
\node[par] (lc) at (9em,10.132em) {};
\draw[>=latex,->] (lc) -- (lab);
\draw (lc) -- (la);
\draw (lc) -- (lb);
\node (llb) at (12em,4.8em) {$n$};
\node[tns] (llc) at (9em,2.868em) {};
\node (lld) at (9em,0.2em) {$np$};
\draw (llc) -- (lld);
\draw (llc) -- (lb);
\draw (llc) -- (llb);
\node (llnp) at (17em,4.8em) {$np$};
\node (llnps) at (23em,4.8em) {$np\ldl s$};
\node[tns] (llc) at (20em,2.868em) {};
\node (lls) at (20em,0.2em) {$s$};
\draw (llc) -- (llnp);
\draw (llc) -- (llnps);
\draw (llc) -- (lls);
\node[tns] (llc2) at (23em,7.668em) {};
\node (lltv) at (20em,9.6em) {$(np\ldl s)\ldr np$};
\node (llnpo) at (26em,9.6em) {$np$};
\draw (llc2) -- (lltv);
\draw (llc2) -- (llnpo);
\draw (llc2) -- (llnps);
\node (llnp) at (29em,4.8em) {$np\ldr n$};
\node (llnps) at (35em,4.8em) {$n$};
\node[tns] (llc) at (32em,2.868em) {};
\node (lls) at (32em,0.2em) {$np$};
\draw (llc) -- (llnp);
\draw (llc) -- (llnps);
\draw (llc) -- (lls);
%
\node (n) at (38em,0.2em) {$n$};
\end{tikzpicture}
\end{center}
\caption{Lexical proof structures for a generalized quantifier noun phrase, a
  transitive verb, a determiner and a noun}
\label{fig:grishin_lex}
\end{figure}

Figure~\ref{fig:grishin_psq} gives, on the left, one of several possible identifications of $n$ and $np$ formulas,
but the only one which produces a proof net with the lexical entries
in the indicate order and the corresponding abstract proof structure
on the right. This abstract proof structure allows us to apply a
contraction directly, as shown in Figure~\ref{fig:grishin_contractions_examples}.

\begin{figure}[h]
\begin{center}
\begin{tikzpicture}
\node (lab) at (9em,9.6em) {$(np\ldr n)\lpr n$};
\node (lb) at (6em,4.8em) {$np\ldr n$};
\node[par] (lc) at (9em,6.886em) {};
\draw[>=latex,->] (lc) -- (lab);
\draw (lc) -- (lb);
\node (llb) at (12em,4.8em) {$n$};
\node[tns] (llc) at (9em,2.868em) {};
\node (lld) at (9em,0.2em) {$np$};
\draw (lc) -- (llb);
\draw (llc) -- (lld);
\draw (llc) -- (lb);
\draw (llc) -- (llb);
\node (llnps) at (19em,0.2em) {$np\ldl s$};
\node[tns] (llc) at (14em,-1.932em) {};
\node (lls) at (14em,-4.6em) {$s$};
\draw (llc) -- (lld);
\draw (llc) -- (llnps);
\draw (llc) -- (lls);
\node[tns] (llc2) at (19em,2.868em) {};
\node (lltv) at (16em,4.8em) {$(np\ldl s)\ldr np$};
\node (llnpo) at (22em,4.8em) {$np$};
\draw (llc2) -- (lltv);
\draw (llc2) -- (llnpo);
\draw (llc2) -- (llnps);
\node (llnp) at (19em,9.6em) {$np\ldr n$};
\node (llnps) at (25em,9.6em) {$n$};
\node[tns] (llc) at (22em,7.668em) {};
\draw (llc) -- (llnp);
\draw (llc) -- (llnps);
\draw (llc) -- (llnpo);
\node (lab) at (33em,9.6em) {$\smash{\apsnode{(np\ldr n)\lpr n}{}}$};
\node (lb) at (30em,4.8em) {$\apsnodei$};
\node[par] (lc) at (33em,6.886em) {};
\draw[>=latex,->] (lc) -- (lab);
\draw (lc) -- (lb);
\node (llb) at (36em,4.8em) {$\apsnodei$};
\node[tns] (llc) at (33em,2.868em) {};
\node (lld) at (33em,0.2em) {$\apsnodei$};
\draw (lc) -- (llb);
\draw (llc) -- (lld);
\draw (llc) -- (lb);
\draw (llc) -- (llb);
\node (llnps) at (43em,0.2em) {$\apsnodei$};
\node[tns] (llc) at (38em,-1.932em) {};
\node (lls) at (38em,-4.6em) {$\smash{\apsnode{}{s}}$};
\draw (llc) -- (lld);
\draw (llc) -- (llnps);
\draw (llc) -- (lls);
\node[tns] (llc2) at (43em,2.868em) {};
\node (lltv) at (40em,4.8em) {$\smash{\apsnode{(np\ldl s)\ldr np}{}}$};
\node (llnpo) at (46em,4.8em) {$\apsnodei$};
\draw (llc2) -- (lltv);
\draw (llc2) -- (llnpo);
\draw (llc2) -- (llnps);
\node (llnp) at (43em,9.6em) {$\smash{\apsnode{np\ldr n}{}}$};
\node (llnps) at (49em,9.6em) {$\smash{\apsnode{n}{}}$};
\node[tns] (llc) at (46em,7.668em) {};
\draw (llc) -- (llnp);
\draw (llc) -- (llnps);
\draw (llc) -- (llnpo);
\end{tikzpicture}
\end{center}
\caption{Judgement $\seq{(np\ldr n)\lpr n,(np\ldl s)\ldr np,np\ldr
n,n}{s}$: proof structure and abstact proof structure}
\label{fig:grishin_psq}
\end{figure}

\begin{figure}[h]
\begin{center}
\begin{tikzpicture}
\node (lab) at (3em,9.6em) {$\smash{\apsnode{(np\ldr n)\lpr n}{}}$};
\node (lb) at (0em,4.8em) {$\apsnodei$};
\node[par] (lc) at (3em,6.886em) {};
\draw[>=latex,->] (lc) -- (lab);
\draw (lc) -- (lb);
\node (llb) at (6em,4.8em) {$\apsnodei$};
\node[tns] (llc) at (3em,2.868em) {};
\node (lld) at (3em,0.2em) {$\apsnodei$};
\draw (lc) -- (llb);
\draw (llc) -- (lld);
\draw (llc) -- (lb);
\draw (llc) -- (llb);
\node (llnps) at (13em,0.2em) {$\apsnodei$};
\node[tns] (llc) at (8em,-1.932em) {};
\node (lls) at (8em,-4.6em) {$\smash{\apsnode{}{s}}$};
\draw (llc) -- (lld);
\draw (llc) -- (llnps);
\draw (llc) -- (lls);
\node[tns] (llc2) at (13em,2.868em) {};
\node (lltv) at (10em,4.8em) {$\smash{\apsnode{(np\ldl s)\ldr np}{}}$};
\node (llnpo) at (16em,4.8em) {$\apsnodei$};
\draw (llc2) -- (lltv);
\draw (llc2) -- (llnpo);
\draw (llc2) -- (llnps);
\node (llnp) at (13em,9.6em) {$\smash{\apsnode{np\ldr n}{}}$};
\node (llnps) at (19em,9.6em) {$\smash{\apsnode{n}{}}$};
\node[tns] (llc) at (16em,7.668em) {};
\draw (llc) -- (llnp);
\draw (llc) -- (llnps);
\draw (llc) -- (llnpo);
%
\node (lld) at (28em,0.2em) {$\smash{\apsnode{(np\ldr n)\lpr n}{}}$};
\node (llnps) at (34em,0.2em) {$\apsnodei$};
\node[tns] (llc) at (31em,-1.932em) {};
\node (lls) at (31em,-4.6em) {$\smash{\apsnode{}{s}}$};
\draw (llc) -- (lld);
\draw (llc) -- (llnps);
\draw (llc) -- (lls);
\node[tns] (llc2) at (34em,2.868em) {};
\node (lltv) at (31em,4.8em) {$\smash{\apsnode{(np\ldl s)\ldr np}{}}$};
\node (llnpo) at (37em,4.8em) {$\apsnodei$};
\draw (llc2) -- (lltv);
\draw (llc2) -- (llnpo);
\draw (llc2) -- (llnps);
\node (llnp) at (34em,9.6em) {$\smash{\apsnode{np\ldr n}{}}$};
\node (llnps) at (40em,9.6em) {$\smash{\apsnode{n}{}}$};
\node[tns] (llc) at (37em,7.668em) {};
\draw (llc) -- (llnp);
\draw (llc) -- (llnps);
\draw (llc) -- (llnpo);
\end{tikzpicture}
\end{center}
\caption{Abstract proof structure and contraction}
\label{fig:grishin_contractions_examples}
\end{figure}
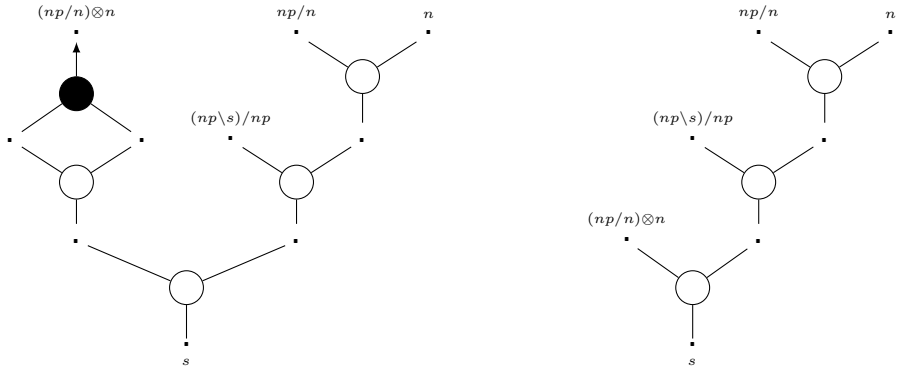

\end{example}

\begin{theorem} A proof structure $P$ is a proof net --- that is, $P$ converts to a tree $T$ --- iff there is a sequent proof of $T$.
\end{theorem}

The proof is an easy adaptation of the proof of \cite{mp}. A detailed
proof can be found in \cite{moot07display}.

By requiring a proof structure to contract to a tree, we actually
compute the structure of the antecedent, which is a pleasant property.

We will look a bit more at the structure of the conversion sequence in
what follows and
the following definition will be useful in this context.

\begin{definition}\label{def:component} Given a proof net $P$, a \emph{component} $C$ of
  $P$ is a maximal subnet of $P$ containing only tensor links.
\end{definition}

From a proof net, we can obtain its components by simply erasing all
cotensor links. The components will be the connected components (in
the graph-theoretic sense) of the resulting graph. In what follows we
will implicitly use the word component to refer only to components
containing at least one tensor link. Though there is no problem in
allowing a component to be a single vertex, the correspondence between
focused sequent proofs and proof nets is more clear when components
are non-trivial.

\subsubsection*{Generalized contractions}

\begin{figure}
\begin{center}
\begin{tikzpicture}
\node (ab) at (3em,9.6em) {$\apsnodei$};
\node (a) at (0,14.4em) {$\smash{\apsnodeh{H}}$};
\node (b) at (6em,14.4em) {$\apsnodei$};
\node[tns] (c) at (3em,12.468em) {};
\draw (c) -- (ab);
\draw (c) -- (a);
\draw (c) -- (b);
\draw [rounded corners,fill=blue!30] (1em,5.1em) rectangle (5em,9.2em) ;
\node (pab) at (3em,4.8em) {$\apsnodei$};
\node (pa) at (0,0) {$\smash{\apsnodec{C}}$};
\node[par] (pc) at (3em,1.732em) {};
\draw (pc) -- (pab);
\path[>=latex,->] (pc) edge (pa);
\draw (b) to [out=50,in=330] (pc);
%
\draw [rounded corners,fill=blue!30] (13em,5.1em) rectangle (17em,9.2em) ;
\node (mb) at (15em,9.5em) {$\smash{\apsnodeh{H}}$};
\node (mb) at (15em,4.7em) {$\smash{\apsnodec{C}}$};
%
\node (rab) at (27em,9.6em) {$\apsnodei$};
\node (ra) at (30em,14.4em) {$\smash{\apsnodeh{H}}$};
\node (rb) at (24em,14.4em) {$\apsnodei$};
\node[tns] (rc) at (27em,12.468em) {};
\draw (rc) -- (rab);
\draw (rc) -- (ra);
\draw (rc) -- (rb);
\draw [rounded corners,fill=blue!30] (25em,5.1em) rectangle (29em,9.2em) ;
\node (rpab) at (27em,4.8em) {$\apsnodei$};
\node (rpa) at (30em,0) {$\smash{\apsnodec{C}}$};
\node[par] (rpc) at (27em,1.732em) {};
\draw (rpc) -- (rpab);
\path[>=latex,->]  (rpc) edge (rpa);
\draw (rb) to [out=130,in=210] (rpc);
\path[ultra thick,>=latex,<-] (12em,7.15em) edge node[above] {$\scriptstyle (G1|G2)^* R\ldr$} (9.2em,7.15em);
\path[ultra thick,>=latex,<-] (18em,7.15em) edge node[above] {$\scriptstyle (G3|G4)^* R\ldl$} (20.8em,7.15em);
\end{tikzpicture}
\end{center}
\caption{Derived Contractions --- Lambek}
\label{fig:lambek_contr_deriv}
\end{figure}
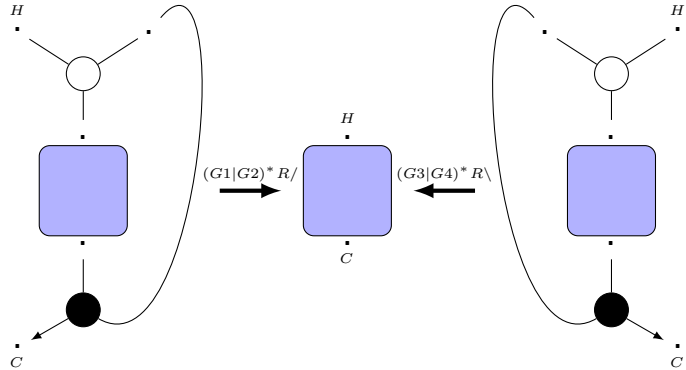

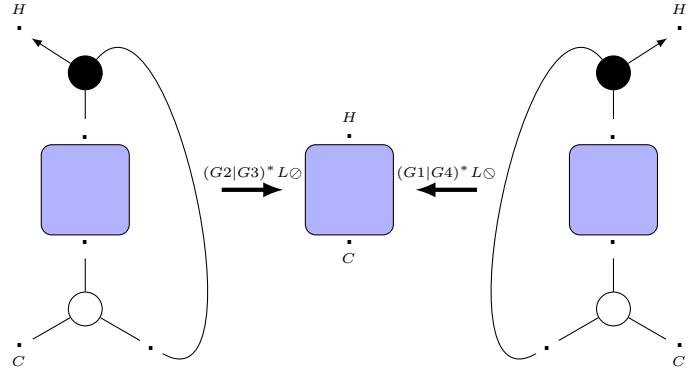
\begin{figure}
\begin{center}
\begin{tikzpicture}
\node (ab) at (3em,9.6em) {$\apsnodei$};
\node (a) at (0,14.4em) {$\smash{\apsnodeh{H}}$};
\node[par] (c) at (3em,12.468em) {};
\draw (c) -- (ab);
\path[>=latex,->] (c) edge (a);
%
\draw [rounded corners,fill=blue!30] (1em,5.1em) rectangle (5em,9.2em) ;
\node (pab) at (3em,4.8em) {$\apsnodei$};
\node (pa) at (0,0) {$\smash{\apsnodec{C}}$};
\node (pb) at (6em,0) {$\apsnodei$};
\node[tns] (pc) at (3em,1.732em) {};
\draw (pc) -- (pab);
\draw (pc) -- (pa);
\draw (pc) -- (pb);
\draw (c) to [out=50,in=330] (pb);
%
\draw [rounded corners,fill=blue!30] (13em,5.1em) rectangle (17em,9.2em) ;
\node (mb) at (15em,9.5em) {$\smash{\apsnodeh{H}}$};
\node (mb) at (15em,4.7em) {$\smash{\apsnodec{C}}$};
%
\node (rab) at (27em,9.6em) {$\apsnodei$};
\node (ra) at (30em,14.4em) {$\smash{\apsnodeh{H}}$};
\node[par] (rc) at (27em,12.468em) {};
\draw (rc) -- (rab);
\path[>=latex,->]  (rc) edge (ra);
%
\draw [rounded corners,fill=blue!30] (25em,5.1em) rectangle (29em,9.2em) ;
\node (rpab) at (27em,4.8em) {$\apsnodei$};
\node (rpa) at (30em,0) {$\smash{\apsnodec{C}}$};
\node (rpd) at (24em,0) {$\apsnodei$};
\node[tns] (rpc) at (27em,1.732em) {};
\draw (rpc) -- (rpab);
\draw (rpc) -- (rpa);
\draw (rpc) -- (rpd);
\draw (rc) to [out=130,in=210] (rpd);
\path[ultra thick,>=latex,<-] (12em,7.15em) edge node[above] {$\scriptstyle (G2|G3)^* L\gdr$} (9.2em,7.15em);
\path[ultra thick,>=latex,<-] (18em,7.15em) edge node[above] {$\scriptstyle (G1|G4)^* L\gdl$} (20.8em,7.15em);
\end{tikzpicture}
\end{center}
\caption{Derived Contractions --- Grishin}
\label{fig:grishin_contr_deriv}
\end{figure}

As can been seen from the figures, the interaction rule introduce
nondeterminism in proof search: a single subtree can be rewritten in
four different trees and this applies recursively for the depth of a
component. However, this is not as bad as it seems:
in many cases, we can ``compile away'' the interaction principles
by permitting contractions in a larger set of configurations than
those shown in Figures~\ref{fig:lambek_contr} and
\ref{fig:grishin_contr}. The contractions for the product and
co-product stay the same, but the contractions for the implications and
co-implications will change as shown in
Figures~\ref{fig:lambek_contr_deriv} and
\ref{fig:grishin_contr_deriv}. In Figure~\ref{fig:lambek_contr_deriv},
the contraction can apply iff there is a path of Grishin tensor
links connecting the two portrayed points above and below the
substructure in the figure. In case this path is empty, the normal
contraction applies and in case this path has length greater than one,
then, by construction, the Lambek tensor link is connected to a
Grishin tensor link, and there is a path from this link through the
displayed substructure. If this path goes left from the first link, we
can apply rule ($G2$) and reduce the distance. If this path goes right
from this first Grishin link, we can apply rule ($G1$) and reduce the
distance as well 
--- in the case of the $R\ldr$
contraction --- or ($G3$) and ($G4$) --- in the case of the $R\ldl$
contraction. 

By up-down symmetry, the contractions of
Figure~\ref{fig:grishin_contr_deriv} require a path of Lambek tensor
connectives with the interaction principles listed. Note that it
suffices to compute one case: the other cases follow from up-down
symmetry and left-to-right symmetry between the interaction principles
and the contractions.

The derived contractions allow us to simplify the reduction sequences
considerably. It is even the case that, whenever the result tree
contains only a single type of constructors (that is, only 
Grishin tensor links or only Lambek tensor links) then we can replace the
interaction principles by the generalized contractions.

\paragraph*{Summary: Proof nets and sequent proofs}

As a useful summary of the correspondence between proof nets and
sequent proofs, we give the following table.

\medskip
\begin{center}
\begin{tabular}{l|ll}
sequent calculus & proof structure & conversion \\ \hline
axiom & axiomatic formula & --- \\
cut & cut formula & --- \\
two-premise rule & tensor link & --- \\
one-premise rule & cotensor link & contraction \\
interaction rule & --- & rewrite \\
\end{tabular}
\end{center}
\medskip

The invertible one-premise rules correspond to both a link and a contraction and
the interaction rules are invisible in the proof structure, appearing
only in the conversion sequence.

With a bit of extra effort in the sequentialization proof --- and the
exclusion of cuts on axioms, because like natural deduction, we cannot
distinguish between the following two sequent proofs

$$
\infer[Cut]{\seq{A}{A}}{\infer[Ax]{\seq{A}{A}}{} & \infer[Ax]{\seq{A}{A}}{}}
\qquad \qquad
\infer[Ax]{\seq{A}{A}}{}
$$

\noindent --- we can show
that these correspondences are 1-on-1, that is each axiomatic formula
in a proof net corresponds to exactly one axiom rule in the sequent
proof, each non-invertible two-premise rule corresponds to exactly one link in the proof
net and each invertible one-premise rule to exactly one link in the proof net and
exactly one contraction in its conversion sequence.

\paragraph*{Discussion} Proof nets provide a solution to the spurious
ambiguity problem of sequent calculus proof search: because of inessential, bureaucratic rule
permutations we can have multiple sequent calculus proofs for what, in essence, corresponds to
the same proof (which corresponds semantically to a
different \emph{reading} of the phrase under consideration). Proof nets, like (product-free) natural deduction, have
different proof objects only for proofs of a judgement which differ
essentially. In addition, the combinatorial possibilities for such readings, which are
obtained by finding a complete matching of the premiss and conclusions
atomic formulas, can easily be enumerated for a given sequence of formulas.

So proof nets have a 1-1 correspondence between proofs and readings,
compute the structure of the sequents, give a graphical representation
which makes the display postulates superfluous and, in certain cases,
can hide the interaction rules by using generalized contractions.


\section{Proof nets and focused display calculus}\label{focusing}
The spurious non-determinism of naive backward-chaining proof search
can be also addressed within the sequent calculus itself, by
introducing an appropriate notion of `normal' derivations.
In \S\ref{lgfocused}, we introduce \textbf{fLG}, a focused version
of the sequent calculus for \LG. In \S\ref{focustonet}, we then
study how to interpret focused derivations from a proof net perspective.

\subsection{fLG: focused display calculus}\label{lgfocused}
The strategy of focusing has been well-studied in the context of linear
logic, starting with the work of Andreoli \cite{focusb}. It is based on the
distinction between \emph{asynchronous} and \emph{synchronous} non-atomic
formulas. The introduction rule for the main connective of an
asynchronous formula is \emph{invertible}; it is non-invertible
for the synchronous formulas. Backward chaining focused proof search starts
with an asynchronous phase where invertible rules are applied 
deterministically until no more candidate formulas remain. At that
point, a non-deterministic choice for a synchronous formula must
be made: this formula is put `in focus', and decomposed in its
subformulae by means of non-invertible rules until no more
non-invertible rules are applicable, at which point one reenters
an asynchronous phase. The main result of \cite{focusb}
is that focused proofs are complete for linear logic.

Focused proof search for the Lambek-Grishin calculus has been studied by
\citeN{bastenhof11polarized} who uses a one-sided presentation of the
calculus. In this section, we implement his focusing regime in the context of
the two-sided sequent format of \cite{bm10cont}. We proceed in 
two steps. First we introduce \textbf{fLG}, the focused version 
of the sequent calculus of \S\ref{simpleseq}. \textbf{fLG} makes
a distinction between focused and unfocused judgements, and has a
set of inference rules to switch between these two. \textbf{fLG} comes
with a term language that is in Curry-Howard correspondence with
its derivations. This term language is a directional refinement of
the \LaMu\  language of \cite{curi:dual00}.

The second step is to give a constructive interpretation for \LG\ derivations
by means of a continuation-passing-style translation:
a mapping $\CPS{\cdot}$ that sends derivations of the multiple-conclusion source logic
to (natural deduction) proofs in a fragment of single-conclusion intuitionistic
Linear Logic MILL (in the categorial terminology: \textbf{LP}). 
For the translation of \cite{bastenhof11polarized} that we follow here,
the target fragment has linear products and negation $A^\perp$, i.e.~a restricted form of linear implication
$A\multimap \perp$, where $\perp$ is a distinguished atomic type, the response type.
Focused source derivations then can be shown to correspond to distinct \emph{normal} natural
deduction proofs in the target calculus.

\[
\textbf{fLG}^{\mathcal{A}}_{/,\otimes,\bs,\oslash,\oplus,\obslash}
\xlongrightarrow{\makebox[.3in]{$\CPS{\cdot}$}}
\textbf{LP}^{\mathcal{A}\cup\{\perp\}}_{\otimes,\cdot^{\perp}}\quad\left(
\xlongrightarrow{\makebox[.3in]{$\CPSlex{\cdot}$}}
\textbf{IL}^{\{e,t\}}_{\times,\Ra}\quad\right)
\]
For the linguistic illustrations in \S\ref{illustrations}, we compose the
CPS translation $\CPS{\cdot}$ with a second mapping $\CPSlex{\cdot}$, that 
establishes the connection with Montague-style semantic representations.
This mapping sends the linear
constructs to their intuitionistic counterparts, and allows \emph{non-linear}
meaning recipes for the translation of the lexical constants.

\subsubsection*{fLG: proofs and terms}\label{terms} We set up \textbf{fLG}
in the Curry-Howard proofs-as-programs fashion, starting from a term language
for which the sequent logic then provides the type system. The term language 
encodes the \emph{logical} steps of a derivation (left and right introduction rules,
and the new set of left and right (de)focusing rules, to be introduced below);
structural rules (residuation, distributivity) leave no trace in the proof terms.

Sequent structures, as in \S\ref{displayseq}, are built out of formulas. Input formulas now are
labeled with variables $x,y,z,\ldots$, output formulas with covariables $\alpha,\beta,\gamma,\ldots$.
To implement the focusing regime, we allow sequents to have one displayed formula \emph{in focus}. 
Writing the focused formula in a box, \textbf{fLG} will have to deal with three types of judgements:
sequents with no formula in focus (we'll call these \emph{structural}),
and sequents with a succedent or antecedent formula in focus. 
\[\nd{X}{Y}\qquad \nD{X}{\,A\,}\qquad \Nd{\,A\,}{Y}\]
Corresponding to the types of sequents, the term language has three types of expressions: 
commands, values and contexts respectively. For commands, we use the metavariables $c,C$,
for values $v,V$, for contexts $e,E$. The typing rules below provide the motivation for
the subclassification.
\begin{equation}\label{rawterms}
\renewcommand{\arraystretch}{1.5}
\begin{array}{c}
v \mathrel{::=} \mu\alpha.C \mid V\quad;\quad
V \mathrel{::=} x \mid v_1\otimes v_2 \mid v\oslash e \mid e\obslash v\\
e \mathrel{::=} \comu x.C \mid E \quad;\quad
E \mathrel{::=} \alpha \mid e_1\oplus e_2 \mid v\bs e \mid e\slash v \\
c \mathrel{::=} \cmdL{x}{E} \mid \cmdR{V}{\alpha}\\
C \mathrel{::=} c \mid \frac{x\  y}{z}.C \mid \frac{x\  \beta}{z}.C \mid \frac{\beta\  x}{z}.C \mid
	\frac{\alpha\ \beta}{\gamma}.C \mid \frac{x\ \beta}{\gamma}.C \mid \frac{\beta\  x}{\gamma}.C \\
\end{array}
\end{equation}
\subsubsection*{Typing rules}\label{ssec:typing_rules}
To enforce the alternation between asynchronous and synchronous phases 
of focused proof search, formulas are associated with a polarity:
\emph{positive} for non-atomic formulas with invertible left introduction
rule: $A\otimes B$, $A\oslash B$, $B\obslash A$;
\emph{negative} for non-atomic formulas with invertible right introduction
rule: $A\oplus B$, $A\bs B$, $B\slash A$. For
atomic formulas, one can fix an arbitrary polarity. Different choices
lead to different prooftheoretic behaviour (and to different interpretations,
once we turn to the CPS translation). We will assume that
atoms are assigned a bias (positive or negative) in the lexicon.
%
Below the typing rules for \textbf{fLG} (restricting attention to
the cut-free system). 

\paragraph*{(Co-)Axiom, (de)focusing}
\[\infer[\textrm{Ax}]{\nD{x:A}{x:A}}{}
\qquad
\infer[\textrm{CoAx}]{\Nd{\alpha:A}{\alpha:A}}{}\]

\[\infer[\mu^*]{\cmdR{V}{\alpha}:(\nd{X}{\alpha:A})}{\nD{X}{V:A}}
\qquad
\infer[\comu^*]{\cmdL{x}{E}:(\nd{x:A}{X})}{\Nd{E:A}{X}}
\]

\[\infer[\comu]{\Nd{\comu x.C:A}{X}}{C:(\nd{x:A}{X})}
\qquad
\infer[\mu]{\nD{X}{\mu\alpha.C:A}}{C:(\nd{X}{\alpha:A})}
\]

First we have the focused version of the axiomatic
sequents, and rules for focusing and defocusing which are new with respect to the
unfocused presentation of \S\ref{simpleseq}. There is a polarity
restriction on the formula $A$ in these rules:
the boxed formula has to be negative for $\textrm{CoAx},\mu, \comu^*$;
for $\textrm{Ax},\comu, \mu^*$ it has to be positive. 
In the \mbox{(Co-)Axiom} cases, $A$ can be
required to be atomic.

From a backward-chaining
perspective, the $\mu,\comu$ rules \emph{remove} the focus from a 
focused succedent or antecedent formula. The result is an unfocused premise
sequent, the domain of applicability of the invertible rules, i.e.~one enters the
asynchronous phase. From the same perspective,
the rules $\mu^*,\comu^*$ place a succedent or antecedent formula in focus, shifting control to
the non-invertible rules of the synchronous phase. The $\mu^*,\comu^*$ rules are in fact
instances of Cut where one of the premises is axiomatic.

\paragraph*{Invertible rules} The term language makes a distinction between
simple commands $c$ (the image of the focusing rules $\comu^*, \mu^*$:
$\cmdL{x}{E}, \cmdR{V}{\alpha}$) from extended commands $C$. The latter start with a sequence of
invertible rewrite rules replacing a logical connective by its structural counterpart.
We impose the requirement that in the asynchronous phase all formulas to which
an invertible rule is applicable are indeed decomposed.

\[\infer[\otimes L]{\frac{x\  y}{z}.C:(\nd{z:A\otimes B}{X})}{C:(\nd{x:A\otimesS y:B}{X})}
\qquad
\infer[\oplus R]{\frac{\alpha\  \beta}{\gamma}.C:(\nd{X}{\gamma:A\oplus B})}{C:(\nd{X}{\alpha:A\oplusS \beta:B})}
\]

\[\infer[\oslash L]{\frac{x\  \beta}{z}.C:(\nd{z:A\oslash B}{X})}{C:(\nd{x:A\oslashS \beta:B}{X})}
\qquad
\infer[\bs R]{\frac{x\  \beta}{\gamma}.C:(\nd{X}{\gamma:A\bs B})}{C:(\nd{X}{x:A\bsS \beta:B})}
\]

\[\infer[\obslash L]{\frac{\beta\  x}{z}.C:(\nd{z:B\obslash A}{X})}{C:(\nd{\beta:B\obslashS x:A}{X})}
\qquad
\infer[\slash R]{\frac{\beta\  x}{\gamma}.C:(\nd{X}{\gamma:B\slash A})}{C:(\nd{X}{\beta:B\slashS x:A})}
\]

\paragraph*{Non-invertible rules} When a positive (negative) formula has been brought into
focus in the succedent (antecedent), one is committed to transfer the focus to its subformulae.

\[\infer[\oplus L]{\Nd{e_1\oplus e_2:B\oplus A}{Y\oplusS X}}{\Nd{e_1:B}{Y} & & \Nd{e_2:A}{X}}
\qquad
\infer[\otimes R]{\nD{X\otimesS Y}{v_1\otimes v_2:A\otimes B}}{\nD{X}{v_1:A} & & \nD{Y}{v_2:B}}
\]

\[\infer[\bs L]{\Nd{v\bs e:A\bs B}{X\bsS Y}}{\nD{X}{v:A} & & \Nd{e:B}{Y}}
\qquad
\infer[\oslash R]{\nD{X\oslashS Y}{v\oslash e:A\oslash B}}{\nD{X}{v:A} & & \Nd{e:B}{Y}}
\]

\[\infer[\slash L]{\Nd{e\slash v:B\slash A}{Y\slashS X}}{\Nd{e:B}{Y} & & \nD{X}{v:A}}
\qquad
\infer[\obslash R]{\nD{Y\obslashS X}{e\obslash v:B\obslash A}}{\Nd{e:B}{Y} & & \nD{X}{v:A}}
\]

\paragraph*{Derived inference rules: focus shifting} To highlight the correspondence with the
algorithm for proof net construction to be discussed in \S\ref{nets}, we will use a
derived rule format for shifting between a conclusion and premise focused formula.
A branch from $(\comu^{*})$ via a sequence (possibly empty) of structural rules 
and rewrite rules to $(\mu)$ is compiled in a derived inference rule with the $\comu^{*}$
restrictions on $A$ and the $\mu$ restrictions on $B$.

\[\begin{array}[t]{c@{\qquad\leadsto\qquad}c}
\infer[\mu]{\nD{X}{\mu\beta.(\div)\cmdL{x}{E}:B}}{\deduce[]{(\div)\cmdL{x}{E}:(\nd{X}{\beta:B)}}{
\deduce[]{\vdots}{\vspace{-.2cm}
\deduce[]{\mathit{(res, distr, rewrite)}}{
\deduce[]{\vdots}{\vspace{-.2cm}
\infer[\comu^{*}]{\cmdL{x}{E}:(\nd{x:A}{Y})}{\Nd{E:A}{Y}}}}}}}
&
\infer[\leftrightharpoons]{\nD{X}{\mu\beta.(\div)\cmdL{x}{E}:B}}{\Nd{E:A}{Y}}
\end{array}\]
For the combinations of $\mu^{*}, \comu^{*}$ and $\mu,\comu$, this results in the
focus shifting rules below. We leave it to the reader to add the terms.

\begin{equation}\label{focusshiftingrules}
\infer[\leftrightharpoons]{\nD{X}{\,B\,}}{\Nd{\,A\,}{Y}}\qquad
\infer[\rightrightharpoons]{\nD{X}{\,B\,}}{\nD{X'}{\,A\,}}\qquad
\infer[\rightleftharpoons]{\Nd{\,B\,}{Y}}{\nD{X}{\,A\,}}\qquad
\infer[\leftleftharpoons]{\Nd{\,B\,}{Y}}{\Nd{\,A\,}{Y'}}
\end{equation}
\subsubsection*{Illustrations} We illustrate the effect of the focusing regime with some alternative ways of
assigning a polarity bias to atomic formulas with a simple Subject-Transitive Verb-Object sentence. Examples
with lexical material filled in would be `everyone seeks/finds a unicorn'.
\begin{equation}\label{likesneeds}
(np_{}^{}/ n_{}^{}\otimes n_{}^{})\cdot\otimes\cdot ((np_{}^{} \bs s_{}^{})/ np_{}^{}\cdot\otimes\cdot (np_{}^{}/ n_{}^{}\cdot\otimes\cdot n_{}^{})) \vdash s_{}^{}
\end{equation}
For the Object we have a Determiner-Noun combination.
For the Subject, we take a product type $(np/n)\otimes n$, so that we have a chance
to illustrate the working of the asynchronous phase of the derivation.
In the unfocused sequent calculus \textbf{sLG}, this sequent has at least seven proofs,
depending on the order of application of the introduction rules for the five
occurrences of the logical connectives involved: $\otimes$ (once), $\slash$ (three times),
$\bs$ (once).

What about the focused calculus \textbf{fLG}? Before answering this question, we have to
decide on the polarization of the atomic types. Suppose we give them uniform negative bias.
There is only one focused proof then: `goal driven', top-down, to use parsing terminology.
In the proof terms, we write \textsf{tv} for the transitive verb; \textsf{det} for the object determiner; 
\textsf{noun} for the object common noun; \textsf{subj} for the subject noun phrase.

\newcommand{\CBVlex}[1]{\widetilde{#1}}
\newcommand{\fboxx}[1]{\framebox{\,\,\rule[-5pt]{0pt}{15pt}#1\,\,}}
\[ \infer[\leftrightharpoons]{(np_{}^{}/ n_{}^{}) \otimes n_{}^{}\cdot\otimes\cdot ((np_{}^{} \bs s_{}^{})/ np_{}^{}\cdot\otimes\cdot (np_{}^{}/ n_{}^{}\cdot\otimes\cdot n_{}^{})) \vdash \fboxx{$s_{}^{}$}}{
 \infer[\slash L]{\fboxx{$(np_{}^{} \bs s_{}^{})/ np_{}^{}$} \vdash ((np_{}^{}/ n_{}^{}\cdot\otimes\cdot n_{}^{})\cdot\bs\cdot s_{}^{})\cdot\slash\cdot (np_{}^{}/ n_{}^{}\cdot\otimes\cdot n_{}^{})}{
 \infer[\bs L]{\fboxx{$np_{}^{} \bs s_{}^{}$} \vdash (np_{}^{}/ n_{}^{}\cdot\otimes\cdot n_{}^{})\cdot\bs\cdot s_{}^{}}{
 \infer[\leftrightharpoons]{np_{}^{}/ n_{}^{}\cdot\otimes\cdot n_{}^{} \vdash \fboxx{$np_{}^{}$}}{
 \infer[\slash L]{\fboxx{$np_{}^{}/ n_{}^{}$} \vdash np_{}^{}\cdot\slash\cdot n_{}^{}}{
\fboxx{$np_{}^{}$} \stackrel{\gamma}{\vdash} np_{}^{} &  \infer[\leftrightharpoons]{n_{}^{} \vdash \fboxx{$n_{}^{}$}}{
\fboxx{$n_{}^{}$} \stackrel{\gamma'}{\vdash} n_{}^{}}}} & \fboxx{$s_{}^{}$} \stackrel{\beta}{\vdash} s_{}^{}} &  \infer[\leftrightharpoons]{np_{}^{}/ n_{}^{}\cdot\otimes\cdot n_{}^{} \vdash \fboxx{$np_{}^{}$}}{
 \infer[\slash L]{\fboxx{$np_{}^{}/ n_{}^{}$} \vdash np_{}^{}\cdot\slash\cdot n_{}^{}}{
\fboxx{$np_{}^{}$} \stackrel{\alpha}{\vdash} np_{}^{} &  \infer[\leftrightharpoons]{n_{}^{} \vdash \fboxx{$n_{}^{}$}}{
\fboxx{$n_{}^{}$} \stackrel{\alpha'}{\vdash} n_{}^{}}}}}}\]
\begin{equation}\label{negativeone}
 \renewcommand{\arraystretch}{1.5}
\begin{array}{c}
\mu \beta_{}.(\displaystyle\frac{y_{}\ z_{}}{\textsf{subj}}.\langle\ \textsf{tv} \upharpoonleft ((Q \mathbin{\bs} \beta_{}) \mathbin{\slash} Q')\ \rangle)\quad\textrm{with}\\
Q:\mu \gamma_{}.\langle\ y_{} \upharpoonleft (\gamma_{} \mathbin{\slash} \mu \gamma'.\langle\ z_{} \upharpoonleft \gamma'\rangle)\rangle\ ,\ 
Q':\mu \alpha_{}.\langle\ \textsf{det} \upharpoonleft (\alpha_{} \mathbin{\slash} \mu \alpha'.\langle\ \textsf{noun} \upharpoonleft \alpha'\rangle)\rangle\\
\end{array}
\end{equation}

As an alternative, suppose basic type $s$ keeps its negative bias, resetting the sentence 
continuation for each clausal domain, but the other basic
types are assigned positive bias. We now have \emph{two} focused derivations: `data driven', bottom-up.
To make sense of this difference, we will have to look at the CPS translation of these proofs, to be introduced
below.
\[ \infer[\leftrightharpoons]{(np_{}^{}/ n_{}^{}) \otimes n_{}^{}\cdot\otimes\cdot ((np_{}^{} \bs s_{}^{-})/ np_{}^{}\cdot\otimes\cdot (np_{}^{}/ n_{}^{}\cdot\otimes\cdot n_{}^{})) \vdash \fboxx{$s_{}^{-}$}}{
 \infer[\slash L]{\fboxx{$np_{}^{}/ n_{}^{}$} \vdash (s_{}^{-}\cdot\slash\cdot ((np_{}^{} \bs s_{}^{-})/ np_{}^{}\cdot\otimes\cdot (np_{}^{}/ n_{}^{}\cdot\otimes\cdot n_{}^{})))\cdot\slash\cdot n_{}^{}}{
 \infer[\leftleftharpoons]{\fboxx{$np_{}^{}$} \vdash s_{}^{-}\cdot\slash\cdot ((np_{}^{} \bs s_{}^{-})/ np_{}^{}\cdot\otimes\cdot (np_{}^{}/ n_{}^{}\cdot\otimes\cdot n_{}^{}))}{
 \infer[\slash L]{\fboxx{$np_{}^{}/ n_{}^{}$} \vdash ((np_{}^{} \bs s_{}^{-})/ np_{}^{}\cdot\bs\cdot (np_{}^{}\cdot\bs\cdot s_{}^{-}))\cdot\slash\cdot n_{}^{}}{
 \infer[\leftleftharpoons]{\fboxx{$np_{}^{}$} \vdash (np_{}^{} \bs s_{}^{-})/ np_{}^{}\cdot\bs\cdot (np_{}^{}\cdot\bs\cdot s_{}^{-})}{
 \infer[\slash L]{\fboxx{$(np_{}^{} \bs s_{}^{-})/ np_{}^{}$} \vdash (np_{}^{}\cdot\bs\cdot s_{}^{-})\cdot\slash\cdot np_{}^{}}{
 \infer[\bs L]{\fboxx{$np_{}^{} \bs s_{}^{-}$} \vdash np_{}^{}\cdot\bs\cdot s_{}^{-}}{
np_{}^{} \stackrel{x_{1}}{\vdash} \fboxx{$np_{}^{}$} & \fboxx{$s_{}^{-}$} \stackrel{\alpha_{0}}{\vdash} s_{}^{-}} & np_{}^{} \stackrel{y_{1}}{\vdash} \fboxx{$np_{}^{}$}}} & n_{}^{} \stackrel{\textsf{noun}}{\vdash} \fboxx{$n_{}^{}$}}} & n_{}^{} \stackrel{z_{0}}{\vdash} \fboxx{$n_{}^{}$}}}\]
\begin{equation}\label{positiveone}
\mu \alpha_{}.(\frac{x'_{}\ z_{}}{\textsf{subj}}.\langle\ x'_{} \upharpoonleft (\comu x_{}.\langle\ \textsf{det} \upharpoonleft (\comu y_{}.\langle\ \textsf{tv} \upharpoonleft ((x_{} \mathbin{\bs} \alpha_{}) \mathbin{\slash} y_{})\ \rangle \mathbin{\slash} \textsf{noun})\ \rangle \mathbin{\slash} z_{})\ \rangle)
\end{equation}

\[ \infer[\leftrightharpoons]{(np_{}^{}/ n_{}^{}) \otimes n_{}^{}\cdot\otimes\cdot ((np_{}^{} \bs s_{}^{-})/ np_{}^{}\cdot\otimes\cdot (np_{}^{}/ n_{}^{}\cdot\otimes\cdot n_{}^{})) \vdash \fboxx{$s_{}^{-}$}}{
 \infer[\slash L]{\fboxx{$np_{}^{}/ n_{}^{}$} \vdash ((np_{}^{} \bs s_{}^{-})/ np_{}^{}\cdot\bs\cdot ((np_{}^{}/ n_{}^{}\cdot\otimes\cdot n_{}^{})\cdot\bs\cdot s_{}^{-}))\cdot\slash\cdot n_{}^{}}{
 \infer[\leftleftharpoons]{\fboxx{$np_{}^{}$} \vdash (np_{}^{} \bs s_{}^{-})/ np_{}^{}\cdot\bs\cdot ((np_{}^{}/ n_{}^{}\cdot\otimes\cdot n_{}^{})\cdot\bs\cdot s_{}^{-})}{
 \infer[\slash L]{\fboxx{$np_{}^{}/ n_{}^{}$} \vdash (s_{}^{-}\cdot\slash\cdot ((np_{}^{} \bs s_{}^{-})/ np_{}^{}\cdot\otimes\cdot np_{}^{}))\cdot\slash\cdot n_{}^{}}{
 \infer[\leftleftharpoons]{\fboxx{$np_{}^{}$} \vdash s_{}^{-}\cdot\slash\cdot ((np_{}^{} \bs s_{}^{-})/ np_{}^{}\cdot\otimes\cdot np_{}^{})}{
 \infer[\slash L]{\fboxx{$(np_{}^{} \bs s_{}^{-})/ np_{}^{}$} \vdash (np_{}^{}\cdot\bs\cdot s_{}^{-})\cdot\slash\cdot np_{}^{}}{
 \infer[\bs L]{\fboxx{$np_{}^{} \bs s_{}^{-}$} \vdash np_{}^{}\cdot\bs\cdot s_{}^{-}}{
np_{}^{} \stackrel{x_{}}{\vdash} \fboxx{$np_{}^{}$} & \fboxx{$s_{}^{-}$} \stackrel{\alpha_{}}{\vdash} s_{}^{-}} & np_{}^{} \stackrel{y_{}}{\vdash} \fboxx{$np_{}^{}$}}} & n_{}^{} \stackrel{z_{}}{\vdash} \fboxx{$n_{}^{}$}}} & n_{}^{} \stackrel{\textsf{noun}}{\vdash} \fboxx{$n_{}^{}$}}}\]
\begin{equation}\label{positivetwo}
\mu \alpha_{}.(\frac{x'_{}\ z_{}}{\textsf{subj}}.\langle\ \textsf{det} \upharpoonleft (\comu y_{}.\langle\ x'_{} \upharpoonleft (\comu x_{}.\langle\ \textsf{tv} \upharpoonleft ((x_{} \mathbin{\bs} \alpha_{}) \mathbin{\slash} y_{})\ \rangle \mathbin{\slash} z_{})\ \rangle \mathbin{\slash} \textsf{noun})\ \rangle)
\end{equation}
\addtolength{\inferLineSkip}{1pt}
\subsubsection*{CPS translation} Let us turn then to the translation that
associates the proofs of the multiple-conclusion source logic \textbf{fLG} with
a constructive interpretation, i.e.~a linear lambda term of the target logic
MILL/\textbf{LP}. CPS translations for \LG\ were introduced in \cite{bernardimm07,bm10cont},
who adapt the call-by-value and call-by-name regimes of \cite{curi:dual00} to
a directional environment. The translation of \cite{bastenhof11polarized}
(following \cite{Gir91}) is an improvement in that it avoids the `administrative redexes'
of the earlier approaches: the image of \LG\ source derivations, under the
 mapping from \cite{bastenhof11polarized} that we present below, are \emph{normal} \textbf{LP} terms.

The target language, on the type level, has the same atoms as the
source language, and in addition a distinguished atom $\perp$, the response type.
Complex types are linear products $-\otimes-$ and a defined negation $A^{\perp}\stackrel{.}{=}A\lolli\perp$.
The CPS translation $\CPS{\cdot}$ maps \textbf{fLG} source types, sequents and their proof terms to
the target types and terms in Curry-Howard correspondence with normal natural deduction
proofs.

\newcommand{\pol}[1]{\textsf{pol}(#1)}

\paragraph*{Types} For positive atoms, $\CPS{p}=p$, for negative atoms
$\CPS{p}=p^{\perp}$. For complex types, the value of $\CPS{\cdot}$
depends on the polarities of the subtypes as shown in Table \ref{cpstrans}.

\begin{table}
  \centering 
  \caption{CPS translation: non-atomic types}\label{cpstrans}
 \renewcommand{\arraystretch}{1.5}
\[\begin{array}{|cc|c|c|c|}\hline
 \pol{A} & \pol{B} & \CPS{A\otimes B} & \CPS{A/B} & \CPS{B\bs A}\\ \hline
 - & - & \CPS{A}^{\perp} \otimes \CPS{B}^{\perp} & \CPS{A} \otimes \CPS{B}^{\perp} & \CPS{B}^{\perp} \otimes \CPS{A}\\ 
 - & + & \CPS{A}^{\perp} \otimes \CPS{B} & \CPS{A} \otimes \CPS{B} & \CPS{B} \otimes \CPS{A}\\ 
 + & - & \CPS{A} \otimes \CPS{B}^{\perp} & \CPS{A}^{\perp} \otimes \CPS{B}^{\perp} & \CPS{B}^{\perp} \otimes \CPS{A}^{\perp}\\ 
 + & + & \CPS{A} \otimes \CPS{B} & \CPS{A}^{\perp} \otimes \CPS{B} & \CPS{B} \otimes \CPS{A}^{\perp}\\ \hline
\end{array}
\]
\[\begin{array}{|cc|c|c|c|}\hline
 \pol{A} & \pol{B} & \CPS{A\oplus B} & \CPS{A\oslash B} & \CPS{B\obslash A}\\ \hline
 - & - & \CPS{A} \otimes \CPS{B} & \CPS{A}^{\perp} \otimes \CPS{B} & \CPS{B} \otimes \CPS{A}^{\perp}\\ 
 - & + & \CPS{A} \otimes \CPS{B}^{\perp} & \CPS{A}^{\perp} \otimes \CPS{B}^{\perp} & \CPS{B}^{\perp} \otimes \CPS{A}^{\perp}\\ 
 + & - & \CPS{A}^{\perp} \otimes \CPS{B} & \CPS{A} \otimes \CPS{B} & \CPS{B} \otimes \CPS{A}\\ 
 + & + & \CPS{A}^{\perp} \otimes \CPS{B}^{\perp} & \CPS{A} \otimes \CPS{B}^{\perp} & \CPS{B}^{\perp} \otimes \CPS{A}\\ \hline
\end{array}
\] 
\end{table}

\paragraph*{Terms} The action of $\CPS{\cdot}$ on terms is given in (\ref{cpsterms}).
We write $\Coco{x},\Coco{\alpha}$ for the target variables
corresponding to source $x,\alpha$. The (de)focusing rules correspond to application/abstraction in the target language. 
Non-invertible (two premise) rules are mapped to linear pair terms; invertible rewrite rules to
the matching deconstructor, the \textbf{case} construct ($\phi,\psi,\xi$ metavariables
for the the (co)variables involved).

\begin{equation}
  \label{cpsterms}
   \renewcommand{\arraystretch}{1.5}
\begin{array}{r@{\qquad}ccc}
\textit{(co)var} & \CPS{x}=\Coco{x} & ; & \CPS{\alpha}=\Coco{\alpha}\\
\textit{linear application} & \CPS{\cmdL{x}{E}}=(\Coco{x}\ \CPS{E}) & ; &
\CPS{\cmdR{V}{\alpha}}=(\Coco{\alpha}\ \CPS{V})\\
\textit{linear abstraction} & \CPS{\comu x.C}=\lambda\Coco{x}.\CPS{C}
& ; & 
\CPS{\mu \alpha.C}=\lambda\Coco{\alpha}.\CPS{C}\\
\textit{linear pair} & \multicolumn{3}{c}{\CPS{\phi\#\psi}=\Zip{\CPS{\phi},\CPS{\psi}}\quad(\#\in\{\otimes,\slash,\bs,\oplus,\oslash,\obslash\})}\\
\textit{case} & \multicolumn{3}{c}{\CPS{\frac{\phi\ \psi}{\xi}.C}=\textbf{case}\ \Coco{\xi}\ \textbf{of}\ \langle\Coco{\phi},\Coco{\psi}\rangle.\CPS{C}}\\
\end{array}
\end{equation}

\paragraph*{Sequents} For sequent hypotheses/conclusions, we have
\begin{equation}\label{cpsatoms}
   \renewcommand{\arraystretch}{1.5}
\begin{array}{c|cc}
\pol{A} & \CPS{x:A} & \CPS{\alpha:A}\\\hline
+ & \Coco{x}:\CPS{A}  & \Coco{\alpha}:\CPS{A}^{\perp} \\
- & \Coco{x}:\CPS{A}^{\perp} &  \Coco{\alpha}:\CPS{A}\\
\end{array}
\end{equation}
Table \ref{cpstrans} then specifies how the translation extends to sequents (replace
logical connectives by their structural counterparts, and target $\otimes$ by the comma for
multiset union).
\renewcommand{\Nd}[2]{\framebox[1.3\width]{$#1$} \vdash #2}
\renewcommand{\nD}[2]{#1 \vdash \framebox[1.3\width]{$#2$}}
\begin{equation}\label{cpstranslation}
\begin{array}{lr@{\quad=\quad}l}
 & \left\lceil C:(\nd{X}{Y})\right\rceil & \CPS{X},\CPS{Y}\vdash_{\textbf{LP}}\CPS{C}:\perp\\
 & \left\lceil\nD{X}{v:A}\right\rceil & \CPS{X}\vdash_{\textbf{LP}}\CPS{v}:\CPS{A}\\
 & \left\lceil\Nd{e:A}{Y}\right\rceil & \CPS{Y}\vdash_{\textbf{LP}}\CPS{e}:\CPS{A}^\perp\\
\end{array}
\end{equation}

\paragraph*{Illustrations} We return to our sample derivations. In  (\ref{cpsandlex})
one finds the CPS image of the source types for transitive verb and determiner under the different
assignments of bias to the atomic subformulas, and the composition with $\CPSlex{\cdot}$, 
assuming $\CPSlex{np}=e$ (entities) and $\CPSlex{s}=\CPSlex{\perp}=t$ (truth values).
For the lexical constants of the illustration, Table \ref{lextrans} gives $\CPSlex{\cdot}$ translations
compatible with the typing. In Table \ref{cpstomg}, these lexical recipes are substituted for
the parameters of the CPS translation.

\begin{equation}
  \label{cpsandlex}
      \renewcommand{\arraystretch}{1.5}
\begin{array}{r@{\quad}c|c|c}
& \textbf{LG} & \CPS{\cdot}^{\perp} & \CPSlex{(\CPS{\cdot}^{\perp})} \\\hline
a. & (np^{+}\bs s^{-})/np^{+} & ((np\otimes s^{\perp})\otimes np)^\perp & ((e\times(tt))\times e)\Ra t\\ 
b. & np^{+}/n^{+} & (np^{\perp}\otimes n)^{\perp} & ((et)\times(et))\Ra t\\
c. & (np^{-}\bs s^{-})/np^{-} & ((np^{\perp\perp}\otimes s^{\perp})\otimes np^{\perp\perp})^\perp & ((((et)t)\times(tt))\times((et)t))\Ra t\\ 
d. & np^{-}/n^{-} & (np^{\perp}\otimes n^{\perp\perp})^{\perp} & ((et)\times(((et)t)t))\Ra t\\
\end{array}
%
\end{equation}

\begin{table}
  \centering 
  \caption{Constants: lexical translations}\label{lextrans}
    \renewcommand{\arraystretch}{1.5}
\[\begin{array}{c@{\quad}r@{\quad}|@{\quad}l}
(np^{+}\bs s^{-})/np^{+} & \W{finds} & \lambda\langle\langle x,c \rangle, y\rangle.(c\  (\textsc{find}^{eet}\ y\ x))\\
(np^{+}/n^{+})\otimes n^{+} & \W{everyone} & \langle \lambda\langle x,y \rangle.(\forall\ \lambda z.(\Rightarrow (y\ z)\ (x\ z))),\textsc{person}^{et}\rangle\\
np^{+}/n^{+} & \W{some} & \lambda\langle x,y \rangle.(\exists\ \lambda z.(\wedge\, (y\ z)\ (x\ z)))\\
n^{+} & \W{unicorn} & \textsc{unicorn}^{et}\\
(np^{-}\bs s^{-})/np^{-} & \W{needs} & \lambda\langle\langle q,c \rangle, q'\rangle.(q\ \lambda x.(\textsc{need}^{((et)t)et}\ q'\ x))\\
(np^{-}/n^{-})\otimes n^{-} & \W{everyone} & \langle \lambda\langle x,w \rangle.(\forall\ \lambda z.(\Rightarrow (w\ \lambda y.(y\ z))\ (x\ z))),
\lambda k.(k\ \textsc{person}^{et})\rangle\\
np^{-}/n^{-} & \W{some} & \lambda\langle x,w \rangle.(\exists\ \lambda z.(\wedge\, (w\ \lambda y.(y\ z))\ (x\ z)))\\
n^{-} & \W{unicorn} & \lambda k.(k\ \textsc{unicorn}^{et})\\
\end{array}
\]
\end{table}
\begin{table}[h]
  \centering 
  \caption{Compositional translations}\label{cpstomg}
  \hrule
  \[\CPS{(\ref{negativeone})}=\lambda \widetilde{\beta}_{}.(\textsf{case}\ \CPSlex{\W{\textsf{subj}}}\ \textsf{of}\ \langle\widetilde{y}_{},\widetilde{z}_{}\rangle.(\CPSlex{\W{\textsf{tv}}} \   \langle  \langle \lambda \widetilde{\gamma}_{}.(\widetilde{y}_{} \   \langle \widetilde{\gamma}_{} , \lambda \widetilde{\gamma}'_{}.(\widetilde{z}_{} \  \widetilde{\gamma}'_{}) \rangle ) , \widetilde{\beta}_{} \rangle  , \lambda \widetilde{\alpha}_{}.(\CPSlex{\W{\textsf{det}}} \   \langle \widetilde{\alpha}_{} , \lambda \widetilde{\alpha}'_{}.(\CPSlex{\W{\textsf{noun}}} \  \widetilde{\alpha}'_{}) \rangle ) \rangle ))\]
\[\CPSlex{\CPS{(\ref{negativeone})}}=\lambda c_{}.(\textsf{$\forall$} \  \lambda x_{}.((\textsf{$\Rightarrow$} \  (\textsf{person} \  x_{})) \  (c_{} \  ((\textsc{needs} \  \lambda w_{}.(\textsf{$\exists$} \  \lambda y_{}.((\textsf{$\wedge$} \  (\textsf{unicorn} \  y_{})) \  (w_{} \  y_{})))) \  x_{}))))\]

\[\CPS{(\ref{positiveone})}=\lambda \widetilde{\alpha}_{}.(\textsf{case}\ \CPSlex{\W{\textsf{subj}}}\ \textsf{of}\ \langle\widetilde{x}'_{},\widetilde{z}_{}\rangle.(\widetilde{x}'_{} \   \langle \lambda \widetilde{x}_{}.(\CPSlex{\W{\textsf{det}}} \   \langle \lambda \widetilde{y}_{}.(\CPSlex{\W{\textsf{tv}}} \   \langle  \langle \widetilde{x}_{} , \widetilde{\alpha}_{} \rangle  , \widetilde{y}_{} \rangle ) , \CPSlex{\W{\textsf{noun}}} \rangle ) , \widetilde{z}_{} \rangle ))\]
\[\CPSlex{\CPS{(\ref{positiveone})}}=\lambda c_{}.(\textsf{$\forall$} \  \lambda x_{}.((\textsf{$\Rightarrow$} \  (\textsc{person} \  x_{})) \  (\textsf{$\exists$} \  \lambda y_{}.((\textsf{$\wedge$} \  (\textsc{unicorn} \  y_{})) \  (c_{} \  ((\textsc{likes} \  y_{}) \  x_{}))))))\]

\[\CPS{(\ref{positivetwo})}=\lambda \widetilde{\alpha}_{}.(\textsf{case}\ \CPSlex{\W{\textsf{subj}}}\ \textsf{of}\ \langle\widetilde{x}'_{},\widetilde{z}_{}\rangle.(\CPSlex{\W{\textsf{det}}} \   \langle \lambda \widetilde{y}_{}.(\widetilde{x}'_{} \   \langle \lambda \widetilde{x}_{}.(\CPSlex{\W{\textsf{tv}}} \   \langle  \langle \widetilde{x}_{} , \widetilde{\alpha}_{} \rangle  , \widetilde{y}_{} \rangle ) , \widetilde{z}_{} \rangle ) , \CPSlex{\W{\textsf{noun}}} \rangle ))\]
\[\CPSlex{\CPS{(\ref{positivetwo})}}=\lambda c_{}.(\textsf{$\exists$} \  \lambda y_{}.((\textsf{$\wedge$} \  (\textsc{unicorn} \  y_{})) \  (\textsf{$\forall$} \  \lambda x_{}.((\textsf{$\Rightarrow$} \  (\textsc{person} \  x_{})) \  (c_{} \  ((\textsc{likes} \  y_{}) \  x_{}))))))\]
\hrule
\end{table}

\subsection{Proof nets and focusing}\label{focustonet}
\newcommand{\semantic}{\textcolor{red}{semantic} }
In this section, we introduce term-labeled proof nets, and
show how a proof term can be read off from the \emph{composition graph}
associated with a net. Our approach is comparable to that of \cite{GR96},
who present an algorithm to compute a linear lambda term from a traversal
of the dynamic graph associated with a proof net for a derivation in the
Lambek calculus. Whereas in the case of the single-conclusion Lambek
calculus, the term associated with a given proof net is unique,
in the case of multiple-conclusion \LG\ there will be the possibility
that the term computation algorithm associates more than one term
with a proof net. These multiple results will then be shown to
correspond to the derivational ambiguity of focused proof search.

\subsubsection*{Reduction tree}
\label{sec:reduction_tree}
When $P$ is a proof net (and therefore
converts to a tensor tree using a sequence $\rho$ of conversions and
contractions) the components of $P$ can bee seen as a parallel
representation of the synchronous phases in
sequent proof search. Taking a closer look at the conversion sequence $\rho$,
we see that all interaction rules operate in one component $C$, the cotensor rules and
the corresponding contractions operate on a component to which it is
attached by both of its active tentacles (i.e.~the tentacles without
the arrow) and the contraction removes a tensor link from this
component. If the main tentacle points to a vertex attached to a
non-trivial component $C'$ then a new component is formed by merging
$C$ (minus the contracted tensor link) and $C'$ into a new
component. When multiple cotensor links have both active tentacles
attached to a single component (Figure~\ref{fig:ex_pnet} shows an
example), we can apply all contractions simultaneously: since a
contraction connects a tensor and a cotensor link at two out of
three tentacles, there cannot be a conflict (multiple cotensor links
connected to a tensor link with \emph{both} contractions being impossible without violating the
definition of proof structures). In addition, when the main vertex of a
cotensor link is the active vertex of another cotensor link, then, if
the other active vertex of this link is connected to the current
component as well, we can apply this contraction immediately. 

So instead of seeing $\rho$ as a \emph{sequence} of reductions, we can
see it as a rooted \emph{tree} of reductions: the initial components
are its leaves (synchronous phases) and the contractions connecting
multiple components to form new components its are branches (the
branches from the active components to their parents correspond to asynchronous
phases) and the final tree --- a single component --- is its root.


\begin{example} Figure~\ref{fig:ex_rooted_tree} shows an example of
  how the view of components given above allows us to see a proof net
  as a tree of components. The shaded subnet boxes are components and
  contain only tensor links. For clarity, the cotensor links are
  shown in the figure as well.

\begin{figure}
\begin{center}
\begin{tikzpicture}
\draw [rounded corners,fill=blue!15] (0,-2) rectangle (1.25,-1);
\node at (0.625,-1.5) {$C_5$};
\draw [rounded corners,fill=blue!40] (0,0) rectangle (1.25,1);
\node at (0.625,0.5) {$C_4$};
\node[minipar] (p1) at (0.625,1.4) {};
\node (a) at (0.2,1.1) {\smash{$\centerdot$}};
\node (b) at (1.05,1.1) {\smash{$\centerdot$}};
\node (c) at (0.625,1.9) {\smash{$\centerdot$}};
\draw (a) -- (p1);
\draw (b) -- (p1);
\draw[->] (p1) -- (c);
\draw [rounded corners,fill=blue!15] (0,2) rectangle (1.25,3);
\node at (0.625,2.5) {$C_3$};
\node[minipar] (p1) at (0.625,3.45) {};
\node (a) at (0.625,3.9) {\smash{$\centerdot$}};
\node (b) at (0.2,7.1) {\smash{$\centerdot$}};
\node (c) at (1.05,3.1) {\smash{$\centerdot$}};
\draw (a) -- (p1);
\draw[->] (p1) -- (c);
\draw (b) to [out=130,in=210] (p1);
\node[minipar] (p1) at (0.625,5.4) {};
\node (a) at (0.2,5.1) {\smash{$\centerdot$}};
\node (b) at (1.05,5.1) {\smash{$\centerdot$}};
\node (c) at (0.625,5.9) {\smash{$\centerdot$}};
\draw (a) -- (p1);
\draw (b) -- (p1);
\draw[->] (p1) -- (c);
%
\node[minipar] (p1) at (0.625,-0.4) {};
\node (a) at (0.2,-0.1) {\smash{$\centerdot$}};
\node (b) at (1.05,-0.1) {\smash{$\centerdot$}};
\node (c) at (0.625,-0.9) {\smash{$\centerdot$}};
\draw (a) -- (p1);
\draw (b) -- (p1);
\draw[->] (p1) -- (c);
\draw [rounded corners,fill=blue!40] (0,4) rectangle (1.25,5);
\node at (0.625,4.5) {$C_2$};
\draw [rounded corners,fill=blue!15] (0,6) rectangle (1.25,7);
\node at (0.625,6.5) {$C_1$};
\draw [rounded corners,fill=blue!15] (3,2) rectangle (4.25,3);
\node at (3.625,2.5) {$C_{3-5}$};
\node[minipar] (p1) at (3.625,3.45) {};
\node (a) at (3.625,3.9) {\smash{$\centerdot$}};
\node (b) at (3.2,5.1) {\smash{$\centerdot$}};
\node (c) at (4.05,3.1) {\smash{$\centerdot$}};
\draw (a) -- (p1);
\draw[->] (p1) -- (c);
\draw (b) to [out=130,in=210] (p1);
\node[minipar] (p1) at (0.625,5.4) {};
\node (a) at (0.2,5.1) {\smash{$\centerdot$}};
\node (b) at (1.05,5.1) {\smash{$\centerdot$}};
\node (c) at (0.625,5.9) {\smash{$\centerdot$}};
\draw (a) -- (p1);
\draw (b) -- (p1);
\draw[->] (p1) -- (c);
%
\node[minipar] (p1) at (0.625,-0.4) {};
\node (a) at (0.2,-0.1) {\smash{$\centerdot$}};
\node (b) at (1.05,-0.1) {\smash{$\centerdot$}};
\node (c) at (0.625,-0.9) {\smash{$\centerdot$}};
\draw (a) -- (p1);
\draw (b) -- (p1);
\draw[->] (p1) -- (c);
\draw [rounded corners,fill=blue!40] (3,4) rectangle (4.25,5);
\node at (3.625,4.5) {$C_{1-2}$};
\draw [rounded corners,fill=blue!40] (6,2) rectangle (7.25,3);
\node at (6.625,2.5) {$C_{1-5}$};
\path[ultra thick,>=latex,->] (1.5,-1.0) edge (2.5,1.0); 
\path[ultra thick,>=latex,->] (1.5,0.75) edge (2.5,1.75); 
\path[ultra thick,>=latex,->] (1.5,2.5) edge (2.5,2.5); 
\path[ultra thick,>=latex,->] (1.5,4.5) edge (2.5,4.5); 
\path[ultra thick,>=latex,->] (1.5,6.5) edge (2.5,5.5); 
\path[ultra thick,>=latex,->] (4.5,2.5) edge (5.5,2.5); 
\path[ultra thick,>=latex,->] (4.5,4.5) edge (5.5,3.5); 
\end{tikzpicture}
\end{center}
\caption{A reduction sequence seen as a rooted tree}
\label{fig:ex_rooted_tree}
\end{figure}
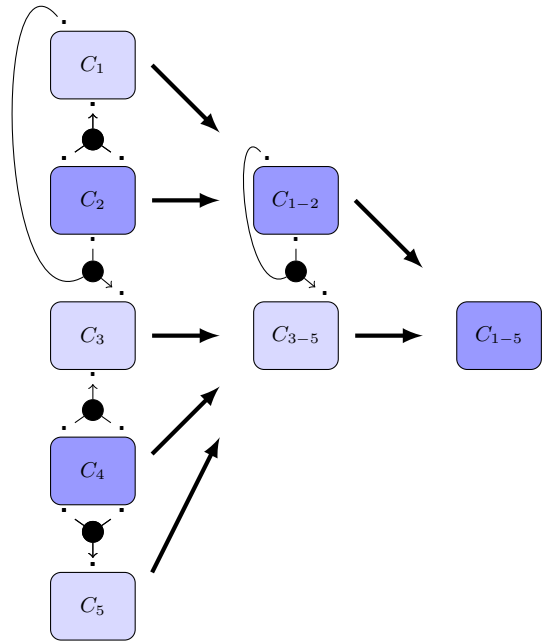

Each interaction rule takes place completely in one of the $C_i$. In
the figure, the components which do not contain the main vertex of a
cotensor link are shown in a darker shade: we will call these
components \emph{active}. In Figure~\ref{fig:ex_rooted_tree}, $C_2$ and
$C_4$ are active. Now, it is easy to show
that whenever there exists a conversions sequence $\rho$, we can
transform it into a conversion sequence $\rho'$ where conversions take
place only in the active components: any conversions in $C_1$ can be
delayed until after the contraction connecting $C_1$ and $C_2$, since
only $C_2$ is relevant for this contraction (it contains both active
vertices of the cotensor link and therefore also the tensor link it
contracts with), and any conversions in $C_5$ can be
delayed until the final component $C_{1-5}$.

In addition, the two active components $C_2$ and $C_4$ are
independent: we can apply conversions to these two components in
parallel.


\end{example}



\subsubsection*{Nets and term labeling}

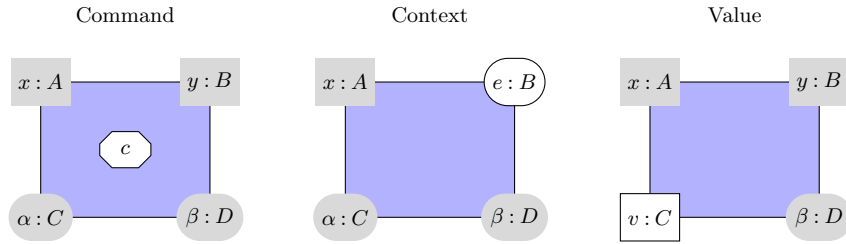
\begin{figure}
\begin{center}
\begin{tikzpicture}

\draw [rounded corners,fill=blue!30] (0,0) rectangle (2.5,2) ;
\node[val,act] at (0,2) {$\labfrm{x}{A}$};
\node[val,act] at (2.5,2) {$\labfrm{y}{B}$};
\node[ctx,act] at (0,0) {$\labfrm{\alpha}{C}$};
\node[ctx,act] at (2.5,0) {$\labfrm{\beta}{D}$};
\node[cmd] at (1.25,1) {$\ c\ $};
\node at (1.25,3) {Command};
\draw [rounded corners,fill=blue!30] (4.5,0) rectangle (7.0,2) ;
\node[val,act] at (4.5,2) {$\labfrm{x}{A}$};
\node[ctx,main] at (7.0,2) {$\labfrm{e}{B}$};
\node[ctx,act] at (4.5,0) {$\labfrm{\alpha}{C}$};
\node[ctx,act] at (7.0,0) {$\labfrm{\beta}{D}$};
\node at (5.75,3) {Context};
\draw [rounded corners,fill=blue!30] (9,0) rectangle (11.5,2) ;
\node[val,act] at (9,2) {$\labfrm{x}{A}$};
\node[val,act] at (11.5,2) {$\labfrm{y}{B}$};
\node[val,main] at (9,0) {$\labfrm{v}{C}$};
\node[ctx,act] at (11.5,0) {$\labfrm{\beta}{D}$};
\node at (10.25,3) {Value};
\end{tikzpicture}
\end{center}
\caption{Proof nets with term labels: commands, context and values}
\label{fig:subnets}
\end{figure}

When assigning a term label to a proof net, we will be interested in
assigning labels to larger and larger subnets of a given proof net,
until we have computed a term for the complete proof net. Like in the
sequent calculus, we distinguish between subnets which are commands,
contexts and values.
Figure~\ref{fig:subnets} shows how we will distinguish these visually: the main formula of a subnet is drawn white,
other formulas are drawn in light gray, values are drawn inside a
rectangle, contexts inside an oval.

\editout{
\begin{definition} The \emph{principal inputs} of a proof structure are its
  hypotheses. The \emph{auxiliary inputs} of a proof net are the minor
  premisses of the asynchronous implication links $R\ldl$ and
  $R\ldr$. The \emph{inputs} of a proof structure are the union of the
  principal and auxiliary inputs.

Symmetrically, the \emph{principal outputs} of a proof structure
are its conclusions, the \emph{auxiliary outputs} are the minor
premisses of the asynchronous co-implication links $L\gdl$ and $L\gdr$ and
the \emph{outputs} are the union of these two sets.
\end{definition}
}

\begin{figure}
{\tikzset{every picture/.style={scale=.85,transform shape, node
      distance=2.5mm}}
\begin{center}
\begin{tabular}{ccc}
\ \\
L$\ldr$ & L$\lpr$ & L $\ldl$ \\
\ \\
\begin{tikzpicture}
\node[val,act] (a) at (2.5,2) {$\labfrm{v}{A}$} ;
\node[ctx,act] (b) at (1.25,0) {$\labfrm{e}{B}$} ;
\node[ctx,main] (ab) at (0,2) {$\labfrm{e\ldr v}{B\ldr A}$} ;
\draw[->] (a) -- (ab);
\draw[->] (b) -- (ab);
\end{tikzpicture}
&
\begin{tikzpicture}
\node[val,pas] (a) at (0,0) {$\labfrm{x}{A}$} ;
\node[val,pas] (b) at (2.5,0) {$\labfrm{y}{B}$} ;
\node[val,pas] (ab) at (1.25,2) {$\labfrm{z}{A\lpr B}$} ;
\node at (1.25,0.8){\fbox{$\displaystyle \frac{x\ y}{z}$}};
\draw[->] (a) -- (ab);
\draw[->] (b) -- (ab);
\end{tikzpicture} 
&
\begin{tikzpicture}
\node[val,act] (a) at (0,2) {$\labfrm{v}{A}$} ;
\node[ctx,act] (b) at (1.25,0) {$\labfrm{e}{B}$} ;
\node[ctx,main] (ab) at (2.5,2) {$\labfrm{v\backsl e}{A\backsl B}$} ;
\draw[->] (a) -- (ab);
\draw[->] (b) -- (ab);
\end{tikzpicture} 
\ \\ 
\ \\ 
R$\ldr$ & R$\lpr$ & R $\ldl$ \\
\ \\

\begin{tikzpicture}
\node[ctx,pas] (ab) at (0,0) {$\labfrm{\gamma}{B \ldr A}$} ;
\node[val,pas] (a) at (2.5,0) {$\labfrm{x}{A}$} ;
\node[ctx,pas] (b) at (1.25,2) {$\labfrm{\beta}{B}$} ;
\node at (1.25,0.8){\fbox{$\displaystyle \frac{\beta x}{\gamma}$}};
\draw[dotted,->] (a) -- (ab);
\draw[->] (b) -- (ab);
\end{tikzpicture}
& 
\begin{tikzpicture}
\node[val,act] (a) at (0,2) {$\labfrm{v_1}{A}$} ;
\node[val,act] (b) at (2.5,2) {$\labfrm{v_2}{B}$} ;
\node[val,main] (ab) at (1.25,0) {$\labfrm{v_1\lpr v_2}{A\lpr B}$} ;
\draw[->] (a) -- (ab);
\draw[->] (b) -- (ab);
\end{tikzpicture} 
&
\begin{tikzpicture}

\node[val,pas] (a) at (0,0) {$\labfrm{x}{A}$} ;
\node[ctx,pas] (b) at (1.5,2) {$\labfrm{\beta}{B}$} ;
\node[ctx,pas] (ab) at (3,0) {$\labfrm{\gamma}{A  \backsl B}$} ;
\node at (1.5,0.8){\fbox{$\displaystyle \frac{x \beta}{\gamma}$}};
\draw[dotted,->] (a) -- (ab);
\draw[->] (b) -- (ab);
\end{tikzpicture}
\\
\ \\
\ \\
L$\gdr$ & L $\gpr$ & L$\gdl$\\
\ \\
\begin{tikzpicture}

\node[ctx,pas] (a) at (2.5,2) {$\labfrm{\alpha}{A}$} ;
\node[val,pas] (b) at (1.25,0) {$\labfrm{x}{B}$} ;
\node[val,pas] (ab) at (0,2) {$\labfrm{z}{B\gdr
A}$} ;
\node at (1.25,1.2){\fbox{$\displaystyle \frac{x \alpha}{z}$}};
\draw[dotted,->] (a) -- (ab);
\draw[->] (b) -- (ab);
\end{tikzpicture} 
&
\begin{tikzpicture}
\node[ctx,act] (a) at (0,0) {$\labfrm{e_1}{A}$} ;
\node[ctx,act] (b) at (2.5,0) {$\labfrm{e_2}{B}$} ;
\node[ctx,main] (ab) at (1.25,2) {$\labfrm{e_1\gpr e_2}{A\gpr B}$} ;
\draw[->] (a) -- (ab);
\draw[->] (b) -- (ab);
\end{tikzpicture} 
&
\begin{tikzpicture}

\node[ctx,pas] (a) at (0,2) {$\labfrm{\alpha}{A}$} ;
\node[val,pas] (b) at (1.25,0) {$\labfrm{x}{B}$} ;
\node[val,pas] (ab) at (2.5,2) {$\labfrm{z}{A\gdl 
B}$} ;
\node at (1.25,1.2){\fbox{$\displaystyle \frac{\alpha x}{z}$}};
\draw[dotted,->] (a) -- (ab);
\draw[->] (b) -- (ab);
\end{tikzpicture}
\\
\ \\
\ \\
R $\gdr$ & R $\gpr$ & R $\gdl$ \\
\ \\

\begin{tikzpicture}

\node[ctx,act] (a) at (2.5,0) {$\labfrm{e}{A}$} ;
\node[val,act] (b) at (1.25,2) {$\labfrm{v}{B}$} ;
\node[val,main] (ab) at (0,0) {$\labfrm{v\gdr e}{B\gdr 
A}$} ;
\draw[->] (a) -- (ab);
\draw[->] (b) -- (ab);
\end{tikzpicture}
&
\begin{tikzpicture}
\node[ctx,pas] (a) at (0,2) {$\labfrm{\alpha}{A}$} ;
\node[ctx,pas] (b) at (2.5,2) {$\labfrm{\beta}{B}$} ;
\node[ctx,pas] (ab) at (1.25,0) {$\labfrm{\gamma}{A\gpr B}$} ;
\node at (1.25,1.2){\fbox{$\displaystyle \frac{\alpha \beta}{\gamma}$}};
\draw[->] (a) -- (ab);
\draw[->] (b) -- (ab);
\end{tikzpicture} 
&
\begin{tikzpicture}

\node[ctx,act] (a) at (0,0) {$\labfrm{e}{A}$} ;
\node[val,act] (b) at (1.25,2) {$\labfrm{v}{B}$} ;
\node[val,main] (ab) at (2.5,0) {$\labfrm{e\gdl v}{A\gdl 
B}$} ;
\draw[->] (a) -- (ab);
\draw[->] (b) -- (ab);
\end{tikzpicture}
\\
\end{tabular}
\end{center}
\caption{\LG\ links with term labeling}
\label{fig:sem_grishin}}
\end{figure}
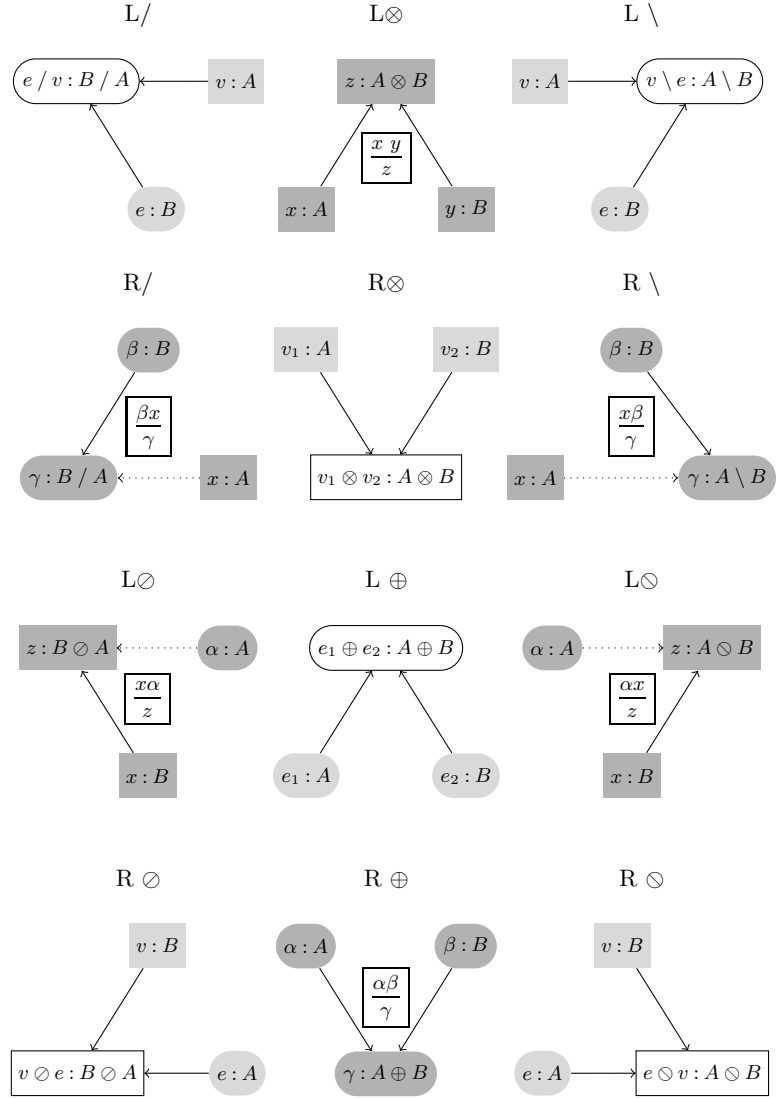

Figure~\ref{fig:sem_grishin}
gives the term-labeled version of the proof net links
corresponding to the logical rules of the sequent calculus.
The flow of information is shown by the arrows: information flow is always from
the active formulas to the main formula of a link, and as a
consequence the complex term can be assigned either to a conclusion or
to a premiss of the link. This is the crucial difference with
term labeling for the single-conclusion Lambek calculus,
where the complex term is always assigned to a conclusion. The cotensor rules, operating on
commands, indicate the prefix for the command corresponding to the term assignment
for the rule (we will see later how commands are formed).

The proof term of an \LG\ derivation is computed on the basis of the
\emph{composition graph} associated with its proof net.

\begin{definition} Given a proof net $P$, the associated 
\emph{composition graph} $cg(P)$ is obtained as follows.
\begin{enumerate}
\item \label{alg:init} all vertices of $P$ with formula label $A$ are expanded into \emph{axiom links}:
edges
  connecting two vertices with formula label $A$;
all links are replaced by the corresponding
links of Figure~\ref{fig:sem_grishin};
\item all vertices in this new structure are
assigned atomic terms of the correct type (variable or covariable) and the
terms for the tensor rules are propagated from the active formulas
to the main formula;
\item\label{alg:subst} all axiom links connecting terms of the same type
(value or context) are collapsed.
\end{enumerate}
\end{definition}

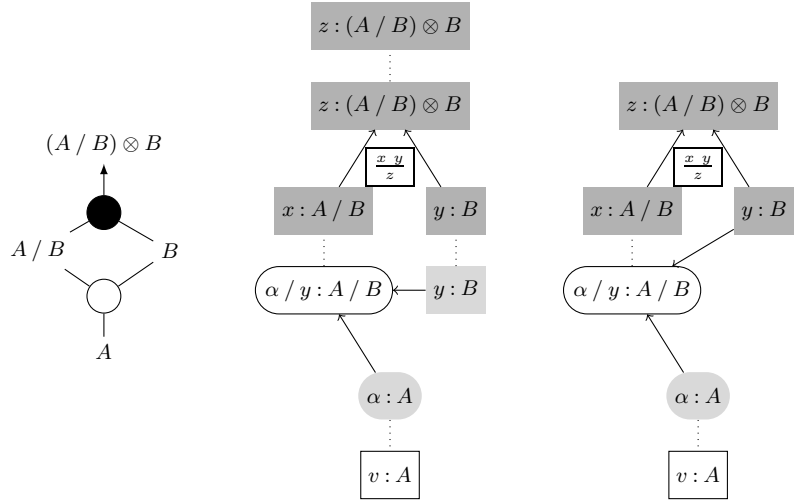
\begin{figure}
\begin{center}
\begin{tikzpicture}
\node (tab) at (0em,-6.4em) {$A$};
\node (ta) at (-3em,-1.8em) {$A\ldr B$};
\node (tb) at (3em,-1.8em) {$B$};
\node[tns] (tc) at (0em,-3.868em) {};
\draw (tc) -- (tab);
\draw (tc) -- (ta);
\draw (tc) -- (tb);
\node (tabx) at (0em,3.0em) {$(A\ldr B)\lpr B$};
\node[par] (tcx) at (0em,-0.068em) {};
\path[>=latex,->]  (tcx) edge (tabx);
\draw (tcx) -- (ta);
\draw (tcx) -- (tb);
\node[ctx,main] (a2) at (10em,-3.6em) {$\labfrm{\alpha\ldr y}{A\ldr B}$} ;
\node[val,act] (b2) at (16em,-3.6em) {$\labfrm{y}{B}$} ;
\node[ctx,act] (b3) at (13em,-8.4em) {$\labfrm{\alpha}{A}$} ;
\node[val,main] (g) at (13em,-12.0em) {$\labfrm{v}{A}$} ;
\draw[->] (b3) -- (a2);
\draw[->] (b2) -- (a2);
\node[val,pas] (a) at (10em,0) {$\labfrm{x}{A\ldr B}$} ;
\node[val,pas] (b) at (16em,0) {$\labfrm{y}{B}$} ;
\node[val,pas] (ab) at (13em,4.8em) {$\labfrm{z}{(A\ldr B)\lpr B}$} ;
\node[val,pas] (abl) at (13em,8.4em) {$\labfrm{z}{(A\ldr B)\lpr B}$} ;
\node at (13em,2.0em){\fbox{$\frac{x\ y}{z}$}};
\draw[->] (a) -- (ab);
\draw[->] (b) -- (ab);
\draw[dotted] (a2) -- (a);
\draw[dotted] (b2) -- (b);
\draw[dotted] (ab) -- (abl);
\draw[dotted] (b3) -- (g);
\node[ctx,main] (a2) at (24em,-3.6em) {$\labfrm{\alpha\ldr y}{A\ldr B}$} ;
\node[ctx,act] (b3) at (27em,-8.4em) {$\labfrm{\alpha}{A}$} ;
\node[val,main] (g) at (27em,-12.0em) {$\labfrm{v}{A}$} ;
\draw[->] (b3) -- (a2);
%
\node[val,pas] (a) at (24em,0) {$\labfrm{x}{A\ldr B}$} ;
\node[val,pas] (b) at (30em,0.0em) {$\labfrm{y}{B}$} ;
\node[val,pas] (ab) at (27em,4.8em) {$\labfrm{z}{(A\ldr B)\lpr B}$} ;
\node at (27em,2.0em){\fbox{$\frac{x\ y}{z}$}};
\draw[->] (a) -- (ab);
\draw[->] (b) -- (ab);
\draw[->] (b) -- (a2);
\draw[dotted] (a2) -- (a);
\draw[dotted] (b3) -- (g);
\end{tikzpicture}
\end{center}
\caption{Proof net (left) and its associated composition graph; in the middle, the expanded net with term annotations;
on the right the result of contracting the substitution links.}
\label{fig:ex_sem}
\end{figure}

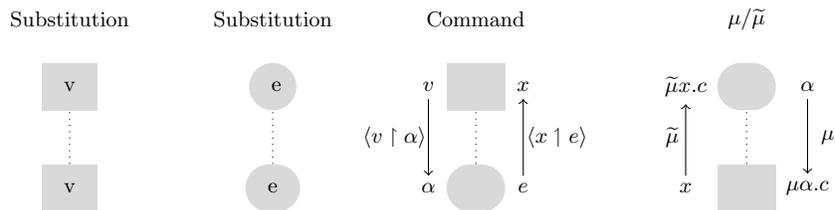
\begin{figure}
\begin{center}
\begin{tikzpicture}

\node (l1) at (0,3.5) {Substitution};
\node[val,act] (a) at (0,1) {\ \  v \ \ };
\node[val,act] (b) at (0,2.5) {\ \ v \ \ };
\draw[dotted] (a) -- (b);
\node (l2) at (3,3.5) {Substitution};
\node[ctx,act] (c) at (3,1) {\ \ e \ \ };
\node[ctx,act] (d) at (3,2.5) {\ \  e\ \ };
\draw[dotted] (c) -- (d);
\node (l3) at (6,3.5) {Command};
\node[ctx,act] (e) at (6,1) {\qquad\quad};
\node[val,act] (f) at (6,2.5) {\qquad\quad};
\draw[dotted] (e) -- (f);
\node (e) at (6.7,1.0) {$e$};
\node (x) at (6.7,2.5) {$x$};
\draw[->] (e) -- (x);
\node (cmdxe) at (7.2,1.75) {$\cmdL{x}{e}$};
\node (a) at (5.3,1.0) {$\alpha$};
\node (v) at (5.3,2.5) {$v$};
\draw[->] (v) -- (a);
\node (cmdva) at (4.8,1.75) {$\cmdR{v}{\alpha}$};
\node (l4) at (10,3.5) {$\mu/\mutilde$};
\node[val,act] (g) at (10,1) {\qquad\quad};
\node[ctx,act] (h) at (10,2.5) {\qquad\quad};
\draw[dotted] (g) -- (h);
\node (muc) at (10.9,1.0) {$\mu \alpha. c$};
\node (al) at (10.9,2.5) {$\alpha$};
\draw[->] (al) -- (muc);
\node (mu) at (11.2,1.75) {$\mu$};
\node (comuc) at (9.1,2.5) {$\comu x. c$};
\node (x) at (9.1,1.0) {$x$};
\draw[->] (x) -- (comuc);
\node (mu) at (8.9,1.75) {$\comu$};
\end{tikzpicture}
\end{center}
\caption{Types of axiom links}
\label{fig:ax_types}
\end{figure}

Figure~\ref{fig:ex_sem} gives an example of the composition graph associated with a net.
In all, the expansion stage gives rise to four types of axiom
links, depending on the type of the term assigned to the $A$ premiss and the
$A$ conclusion. These cases are summarized in Figure~\ref{fig:ax_types}.
The substitution links are collapsed in the final stage of the construction of
the composition graph; the command and $\mu/\comu$ cases are the ones that
remain. 

Given the composition graph $cg(P)$ associated with a proof net $P$, we compute
terms for it as follows.

\begin{enumerate}
\item\label{alg:subnet} we compute all maximal subnets of $cg(P)$,
  which consist of a set of tensor links with a single main formula,
  marking all these links as visited;
\item \label{alg:iter} while $cg(P)$ contains unvisited links do the following:
\begin{enumerate}
\item \label{alg:cmd} follow an unvisited command link attached to a
  previously calculated maximal subnet, forming a correct command
  subnet; like before, we restrict to \emph{active} subnets which do
  not contain (or allow us to reach through an axiom) the main formula of a negative link;
\item \label{alg:neg} for each negative link with both active formulas attached to the current
  command subnet, pass to the main formula of the negative link,
  forming a new command, repeat this step until no such negative
  links remain attached;
\item \label{alg:mu} follow a $\mu$ or $\comu$ link to a new vertex, forming a larger value or context
  subnet and replacing the variable previously assigned to the newly
  visited vertex by the $\mu$ value or $\comu$ context.
\end{enumerate}
\end{enumerate}

The algorithm 
stays quite close to the focused proof nets of
the previous section: the maximal subnets of
step~\ref{alg:subnet} are \emph{rooted} versions of the components we
have used before, with the directions of the arrows potentially
splitting components into multiple rooted components
(Figure~\ref{fig:semb} will give an example) and the asynchronous phases,
which consisted of one or more contractions for cotensor links,
will now consist of a passage through a
command link, followed by zero or more cotensor links, followed by either a
$\mu$ or a $\comu$ link, the result being a new, larger subnet.
The term assignment algorithm is a way to enumerate the
non-equivalent proof terms of a net. Given that these
terms are isomorphic to focused sequent proofs, it is no coincidence that the computation of
the proof terms looks a lot like the sequentialisation algorithm.\footnote{The connection between proof net sequentialisation and focusing for linear logic is
explored in \cite{am99proofnets}} The following lemma is easy to prove.

\begin{lemma} If $P$ is a proof net (with a pairing of command and
  $\mu/\comu$ links) and $v$ is a term calculated for $P$ using this
  pairing then there is a sequent proof $\pi$ which is assigned $v$ as well.
\end{lemma}

This lemma is easily proved by induction on the depth of the tree: it
holds trivially for the leaves (which are rooted components), and, inductively, each command, contensor,
$\mu/\comu$ sequence will produce a sequent proof of the same term: in
fact each such step corresponds exactly to the derived inference rules
for focus shifting discussed in \S\ref{ssec:typing_rules}.

To summarize: the difference between computing terms for
proof nets in the Lambek calculus and in \LG\ can be characterized by the
following statements:

\begin{description}
\item{Lambek calculus:} the (potential) terms are given through a bijection between premiss and conclusion atomic
formulas (ie.\ a complete matching of the axioms),
\item{\LG:} the (potential) terms are given through a
  bijection between premiss and conclusion atomic formulas \emph{plus} a
bijection between command and $\mu/\comu$ axioms.
\end{description}

We speak of
  \emph{potential} terms, since in the case of the Lambek calculus
  only proof nets can be assigned a term, whereas in the \LG\ case we
  need proof nets plus a coherent bijection between command and
  $\mu/\comu$ axioms, where the $\mu$ or $\comu$ rule is applied to
  one of the free variables of the command $c$.
  
\subsubsection*{Illustrations}
\label{illustrations}
\begin{figure}
\begin{center}
\begin{tikzpicture}
\draw [rounded corners,fill=blue!30] (10em,-3.6em) -- (13em,-3.6em) --
(13em,0.0em) -- (16em,0.0em) -- (16em,-8.4em) -- (10em,-8.4em) -- cycle;
\node[ctx,main] (a2) at (10em,-3.6em) {$\labfrm{\alpha\ldr y}{A\ldr B}$} ;
\node[ctx,act] (b3) at (13em,-8.4em) {$\labfrm{\alpha}{A}$} ;
\node[val,main] (g) at (13em,-12.0em) {$\labfrm{v}{A}$} ;
\node[val,pas] (a) at (10em,0) {$\labfrm{x}{A\ldr B}$} ;
\node[val,pas] (b) at (16em,0.0em) {$\labfrm{y}{B}$} ;
\node[val,pas] (ab) at (13em,4.8em) {$\labfrm{z}{(A\ldr B)\lpr B}$} ;
\node at (13em,2.0em){\fbox{$\frac{x\ y}{z}$}};
\draw[->] (a) -- (ab);
\draw[->] (b) -- (ab);
\draw[dotted] (a2) -- (a);
\draw[dotted] (b3) -- (g);
\draw [rounded corners,fill=blue!30] (24em,0.0em) -- (30em,0.0em) -- (30em,-8.4em) -- (24em,-8.4em) -- cycle;
\node[cmd] (cmd) at (27em,-4.8em) {$\cmdL{x}{\alpha\ldr y}$};
\node[ctx,act] (b3) at (27em,-8.4em) {$\labfrm{\alpha}{A}$} ;
\node[val,main] (g) at (27em,-12.0em) {$\labfrm{v}{A}$} ;
\node[val,pas] (a) at (24em,0) {$\labfrm{x}{A\ldr B}$} ;
\node[val,pas] (b) at (30em,0.0em) {$\labfrm{y}{B}$} ;
\node[val,pas] (ab) at (27em,4.8em) {$\labfrm{z}{(A\ldr B)\lpr B}$} ;
\node at (27em,2.0em){\fbox{$\frac{x\ y}{z}$}};
\draw[->] (a) -- (ab);
\draw[->] (b) -- (ab);
\draw[dotted] (b3) -- (g);
\draw [rounded corners,fill=blue!30] (38em,0.0em) -- (44em,0.0em) -- (44em,-8.4em) -- (38em,-8.4em) -- cycle;
\node[cmd] (cmd) at (41em,-4.8em) {$\frac{x\ y}{z}\cmdL{x}{\alpha\ldr y}$};
\node[ctx,act] (b3) at (41em,-8.4em) {$\labfrm{\alpha}{A}$} ;
\node[val,main] (g) at (41em,-12.0em) {$\labfrm{v}{A}$} ;
%
\node[val,pas] (ab) at (41em,0.0em) {$\labfrm{z}{(A\ldr B)\lpr B}$} ;
\draw[dotted] (b3) -- (g);
\end{tikzpicture}
\end{center}
\caption{Computing the proof term from a composition graph}
\label{fig:ex_compute_sem}
\end{figure}
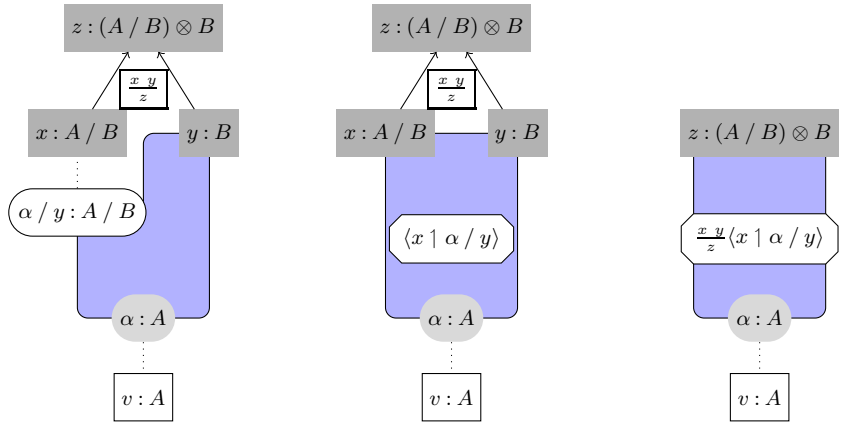

Figure~\ref{fig:ex_compute_sem} shows how to compute the term for the
example proof net of Figure~\ref{fig:ex_sem}, starting from the composition
graph (on the right).
We first look for the components
(step~\ref{alg:subnet}). Since there is only a single tensor link,
this is simple. Figure~\ref{fig:ex_compute_sem} shows, on the left, the
context subnet corresponding to this link.

Now, there is only one command to follow from here (step~\ref{alg:cmd}), which produces the
command shown in the middle of
Figure~\ref{fig:ex_compute_sem}. Applying the cotensor link
(step~\ref{alg:neg}) produces the figure shown on the right. The final
$\mu$ link (step~\ref{alg:mu}, not shown) produces the completed term for this
proof net. 

$$v = \mu \alpha. \frac{x\ y}{z}\cmdL{x}{\alpha\ldr y}$$ 

Some remarks about this example. First, some of the axioms can be
traversed in only one of the two possible directions: in cut-free
proof nets, command links move either towards the active formulas of cotensor links
or towards ``dead ends'': hypotheses or conclusions of the proof
net. And since we want to compute the value of $v$ for the example
proof net, it only makes sense to apply a $\mu$ rule to compute this
value: we always ``exit'' the proof net from a designated conclusion.
With a slight modification to the algorithm that reads off terms
from a composition graph,  we could also compute \emph{commands} for proof nets,
or compute the \emph{context} for a designated \emph{premiss} of the net.


\newcommand{\lexuni}{\textsf{noun}}
\newcommand{\lexseeks}{\textsf{tv}}
\newcommand{\lexdet}{\textsf{det}}
\newcommand{\lexsuj}{\textsf{subj}}

\begin{figure}
\begin{center}
\begin{tikzpicture}[scale=.9]
\node[ctx,act] (vp2a) at (2.5,1.5) {$\labfrm{x\backslash \beta}{np\ldl s}$} ;
\node[ctx,act] (npx) at (0,2.5) {$\labfrm{\gamma}{np}$};
\node[val,act] (gq2) at (0,1.5) {$\labfrm{x}{np}$}; 
\node[ctx,act] (det) at (-1.0,4.0) {$\labfrm{\gamma/z'}{np\ldr n}$};
\node[val,pas] (det2) at (-1.0,5.0) {$\labfrm{y'}{np\ldr n}$};
\node[val,main] (gqp) at (0,6.5) {$\labfrm{\lexsuj}{(np\ldr n)\lpr n}$};
\node[val,pas] (n) at (1.5,4.5) {$\labfrm{z'}{n}$};
\node[ctx,act] (tv) at (0.75,8.0) {$\labfrm{(x\backslash \beta)/y}{(np 
\backsl s)/np}$} ;
\node[val,main] (tvlex) at (0.75,9.0) {$\labfrm{\lexseeks}{(np 
\backsl s)/np}$} ;
\node[val,act] (auni) at (3.375,8.0) {$\labfrm{y}{np}$};
\node[ctx,act] (aunib) at (3.375,9) {$\labfrm{\alpha}{np}$};
\node[ctx,act] (some) at (2.25,10.25) {$\labfrm{\alpha/\lexuni}{np/n}$}; 
\node[val,main] (somelex) at (2.25,11.25) {$\labfrm{\lexdet}{np/n}$};
\node[val,main] (uniclex) at (4.75,10.25) {$\labfrm{\lexuni}{n}$};
\node[ctx,act] (sf) at (1.25,0.0) {$\labfrm{\beta}{s}$};
\node[val,main] (sg) at (1.25,-1.0) {$\labfrm{x'}{s}$};

\draw[dotted] (det) -- (det2);
\draw[->] (n) -- (det);
\draw[<-] (gqp) -- (n);
\draw[<-] (gqp) -- (det2);
%
%
\draw[->] (uniclex) -- (some);
\draw[->] (aunib) -- (some);
\draw[dotted] (npx) -- (gq2);
\draw[->] (npx) -- (det);
\draw[->] (auni) -- (tv);
\draw[dotted] (sf) -- (sg);
\draw[dotted] (some) to (somelex);
\draw[dotted] (tv) -- (tvlex);
\draw[dotted] (auni) -- (aunib);
\draw[->] (sf) -- (vp2a);
\draw[->] (gq2) -- (vp2a);
\draw[->] (vp2a) .. controls (3.5,3.5) and (3.5,6.5) .. (tv);
\draw [rounded corners,fill=blue!30] (10.25,10.25) -- (12.75,10.25) -- (12.75,9) --
(10.25,9) -- cycle;
\draw [rounded corners,fill=blue!30] (8.75,8.0) -- (11.375,8.0) --
(11.375,0.0) -- (8,0.0) -- (8,1.5) -- (10,1.5) -- (10,7.0) -- (8.75,7.0)  -- cycle;
\draw [rounded corners,fill=blue!30] (7.0,4.0) -- (8.0,4.0) -- (8.0,4.5)
-- (9.0,4.5) -- (9.0,2.5) -- (7.0,2.5) -- cycle;
\node at (8,3.25) {$C_1$};
\node at (10.7,4.25) {$C_2$};
\node at (11.375,9.625) {$C_3$};
%
\node[ctx,act] (npx) at (8,2.5) {$\labfrm{\gamma}{np}$};
\node[val,act] (gq2) at (8,1.5) {$\labfrm{x}{np}$}; 
\node[ctx,act] (det) at (7.0,4.0) {$\labfrm{\gamma/z'}{np\ldr n}$};
\node[val,pas] (det2) at (7.0,5.0) {$\labfrm{y'}{np\ldr n}$};
\node[val,main] (gqp) at (8,6.5) {$\labfrm{\lexsuj}{(np\ldr n)\lpr n}$};
\node[val,pas] (n) at (9.0,4.5) {$\labfrm{z'}{n}$};
\node[ctx,act] (tv) at (8.75,8.0) {$\labfrm{(x\backslash \beta)/y}{(np 
\backsl s)/np}$} ;
\node[val,main] (tvlex) at (8.75,9.0) {$\labfrm{\lexseeks}{(np 
\backsl s)/np}$} ;
\node[val,act] (auni) at (11.375,8.0) {$\labfrm{y}{np}$};
\node[ctx,act] (aunib) at (11.375,9) {$\labfrm{\alpha}{np}$};
\node[ctx,act] (some) at (10.25,10.25) {$\labfrm{\alpha/\lexuni}{np/n}$}; 
\node[val,main] (somelex) at (10.25,11.25) {$\labfrm{\lexdet}{np/n}$};
\node[val,main] (uniclex) at (12.75,10.25) {$\labfrm{\lexuni}{n}$};
\node[ctx,act] (sf) at (9.25,0.0) {$\labfrm{\beta}{s}$};
\node[val,main] (sg) at (9.25,-1.0) {$\labfrm{x'}{s}$};

\path[dotted] (sf) edge [left] node {$\scriptstyle (\mu 1)$} (sg);
\path[dotted] (gq2) edge [left] node {$\scriptstyle (\mu 2)$} (npx);
\path[dotted] (auni) edge [left] node {$\scriptstyle (\mu 3)$} (aunib);
\path[dotted] (det) edge [left] node {$\scriptstyle (c 1)$} (det2);
\path[dotted] (some) edge [left] node {$\scriptstyle (c 3)$} (somelex);
\path[dotted] (tv) edge [left] node {$\scriptstyle (c 2)$} (tvlex);
\draw[->] (det2) -- (gqp);
\draw[->] (n) -- (gqp);
\end{tikzpicture}
\end{center}
\caption{Composition graph (left) and initial components (right) for the ``\lexsuj{} \lexseeks{}
  \lexdet{} \lexuni{}'' example}
\label{fig:semb}
\end{figure}
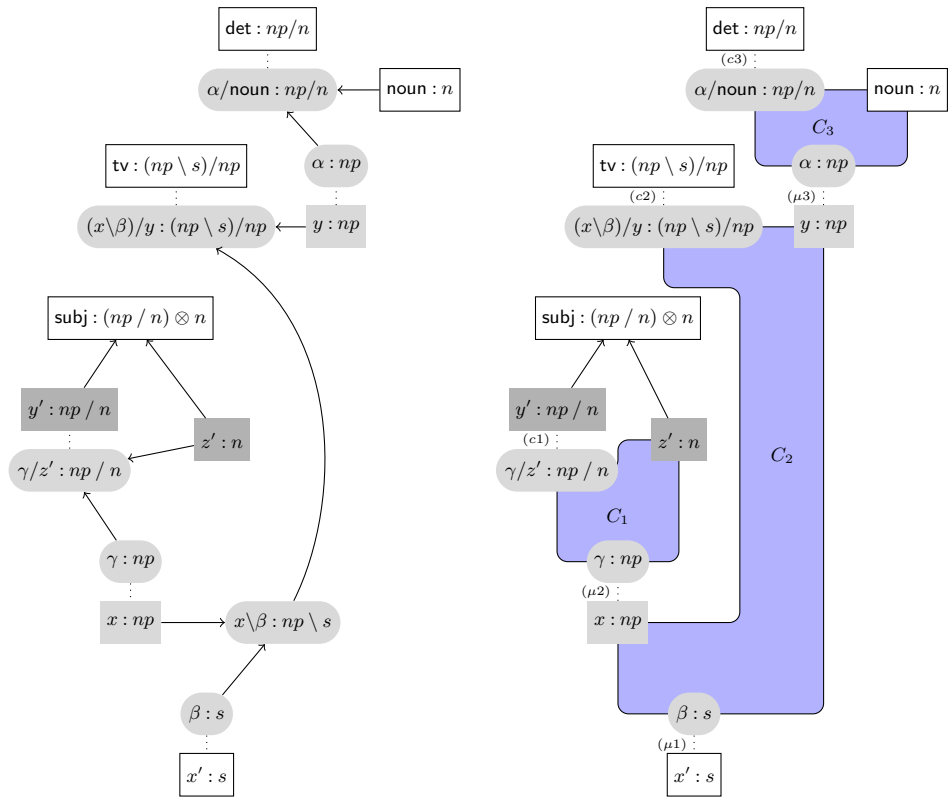

Figure~\ref{fig:semb} returns to our ``\lexsuj{} \lexseeks{}
  \lexdet{} \lexuni{}'' example.  On the left we see the composition graph for the example  of Figure~\ref{fig:grishin_psq}.

The only cotensor link in the figure has the node $\labfrm{\lexsuj}{(np\ldr n)\lpr n}$
as its main formula.
When we compute the rooted components, we
see that there are three, shown on the right of the figure.
%
There are three command axioms, one for the root node of each of the three
components, $C_1$ to $C_3$ on the right hand side of the figure; these are numbered $c1$ to $c3$ next to
the corresponding links with the same number as the corresponding component. There are also three $\mu/\comu$ links (numbered $\mu 1$
to $\mu 3$).  Figure~\ref{fig:ex_readingb} gives a schematic
representation of the proof net of Figure~\ref{fig:semb}.

Since we are interested in calculating the value of $x'$, $(\mu 1)$ will
be the last link we pass in the proof net and therefore we will pass
it downwards, producing a term of the form $\mu \beta. c$.  Figure~\ref{fig:ex_readingb} gives a schematic
representation of the proof net of Figure~\ref{fig:semb}. The arrows
next to the $\mu/\comu$ links indicate the different possibilities for
traversing the link and whether this traversal corresponds to a $\mu$ or
a $\comu$ link. 




\begin{figure}
\begin{center}
\begin{tikzpicture}
\draw [rounded corners,fill=blue!40] (0,2) rectangle (1.25,3);
\node at (0.625,2.5) {$C_1$};
\node (a) at (1.35,4.5) {\smash{$\centerdot$}};
\node (b) at (2.0,4.5) {\smash{$\centerdot$}};
\node (at) at (1.35,4.65) {};
\node (bt) at (2.0,4.65) {};
\draw[->] (at) -- (bt);
\node (cmd2) at  (1.625,4.9) {\smash{$\mu_1$}};
\draw[dotted] (a) -- (b);
\node (a) at  (0.625,3.1) {\smash{$\centerdot$}};
\node (b) at  (0.625,3.9) {\smash{$\centerdot$}};
\draw[dotted] (a) -- (b);
\node (ar) at  (0.725,3.10) {};
\node (br) at  (0.725,3.90) {};
\draw[->] (br) -- (ar);
\node (al) at  (0.525,3.10) {};
\node (bl) at  (0.525,3.90) {};
\draw[->] (al) -- (bl);
\node (mu3) at (0.95,3.55){$\comu_2$};
\node (comu3) at (0.30,3.5){$\mu_2$};
\node (a) at  (0.625,5.1) {\smash{$\centerdot$}};
\node (b) at  (0.625,5.9) {\smash{$\centerdot$}};
\draw[dotted] (a) -- (b);
\node (br) at  (0.525,5.10) {};
\node (ar) at  (0.525,5.90) {};
\draw[->] (ar) -- (br);
\node (bl) at  (0.725,5.10) {};
\node (al) at  (0.725,5.90) {};
\draw[->] (bl) -- (al);
\node (mu3) at (0.30,5.5){$\mu_3$};
\node (comu3) at (0.95,5.55){$\comu_3$};
\node (a) at  (-0.1,2.5) {\smash{$\centerdot$}};
\node (b) at  (-0.8,2.5) {\smash{$\centerdot$}};
\node (cmd1) at  (-0.45,2.75) {\smash{$c_1$}};
\draw[->] (a) -- (b);
\node (a) at  (-0.1,4.5) {\smash{$\centerdot$}};
\node (b) at  (-0.8,4.5) {\smash{$\centerdot$}};
\node (cmd2) at  (-0.45,4.75) {\smash{$c_2$}};
\draw[->] (a) -- (b);
\node (a) at  (-0.1,6.5) {\smash{$\centerdot$}};
\node (b) at  (-0.8,6.5) {\smash{$\centerdot$}};
\node (cmd1) at  (-0.45,6.75) {\smash{$c_3$}};
\draw[->] (a) -- (b);
\draw [rounded corners,fill=blue!40] (0,4) rectangle (1.25,5);
\node at (0.625,4.5) {$C_2$};
\draw [rounded corners,fill=blue!40] (0,6) rectangle (1.25,7);
\node at (0.625,6.5) {$C_3$};
\draw [rounded corners,fill=blue!40] (5,2) rectangle (6.25,3);
\node at (5.625,2.5) {$C_{1}$};
\draw [rounded corners,fill=blue!40] (5,4) rectangle (6.25,5);
\node at (5.625,4.5) {$C_{2-3}$};
\node (a) at (6.35,4.5) {\smash{$\centerdot$}};
\node (b) at (7.0,4.5) {\smash{$\centerdot$}};
\draw[dotted] (a) -- (b);
\node (at) at (6.35,4.65) {};
\node (bt) at (7.0,4.65) {};
\draw[->] (at) -- (bt);
\node (cmd2) at  (6.625,4.9) {\smash{$\mu_1$}};
\node (a) at  (5.625,3.1) {\smash{$\centerdot$}};
\node (b) at  (5.625,3.9) {\smash{$\centerdot$}};
\draw[dotted] (a) -- (b);
\node (ar) at  (5.725,3.10) {};
\node (br) at  (5.725,3.90) {};
\draw[->] (br) -- (ar);
\node (al) at  (5.525,3.10) {};
\node (bl) at  (5.525,3.90) {};
\draw[->] (al) -- (bl);
\node (mu3) at (5.95,3.55){$\comu_2$};
\node (comu3) at (5.30,3.5){$\mu_2$};
\node (a) at  (4.9,2.5) {\smash{$\centerdot$}};
\node (b) at  (4.2,2.5) {\smash{$\centerdot$}};
\node (cmd1) at  (4.55,2.75) {\smash{$c_1$}};
\draw[->] (a) -- (b);
\node (a) at  (4.9,4.5) {\smash{$\centerdot$}};
\node (b) at  (4.2,4.5) {\smash{$\centerdot$}};
\node (cmd2) at  (4.55,4.75) {\smash{$c_3$}};
\draw[->] (a) -- (b);
\draw [rounded corners,fill=blue!40] (10,2) rectangle (11.25,3);
\node at (10.625,2.5) {$C_{1-3}$};
\node (a) at  (9.9,2.5) {\smash{$\centerdot$}};
\node (b) at  (9.2,2.5) {\smash{$\centerdot$}};
\node (cmd2) at  (9.55,2.75) {\smash{$c_1$}};
\draw[->] (a) -- (b);
\node (a) at (11.35,2.5) {\smash{$\centerdot$}};
\node (b) at (12.0,2.5) {\smash{$\centerdot$}};
\draw[dotted] (a) -- (b);
\node (at) at (11.35,2.65) {};
\node (bt) at (12.0,2.65) {};
\draw[->] (at) -- (bt);
\node (cmd2) at  (11.625,2.9) {\smash{$\mu_1$}};
\path[ultra thick,>=latex,->] (2.5,2.5) edge (3.7,2.5); 
\path[ultra thick,>=latex,->] (2.5,4.5) edge (3.7,4.5); 
\path[ultra thick,>=latex,->] (2.5,6.5) edge (3.7,5.5); 
\path[ultra thick,>=latex,->] (7.5,2.5) edge (8.7,2.5); 
\path[ultra thick,>=latex,->] (7.5,4.5) edge (8.7,3.5); 
\end{tikzpicture}
\end{center}
\caption{Matching: $c_2-\comu_3$,
$c_3-\comu_2$, $c_2-\mu_1$. Reading: $\lexsuj < \lexdet < \lexseeks$.}
\label{fig:ex_readingb}
\end{figure}

If both $np$ arguments of the transitive verbs are
lexically assigned a positive bias, then we can only pass the two axioms
$\mu_2/\comu_2$ and $\mu_3/\comu_3$ in the $\comu_2$ and $\comu_3$
directions, following the arrows away from component $C_2$. This will
necessarily mean that the first command is $c_2$ and that we can
follow this command either with $\comu_2$ 
(going to
the component of the subject and producing the narrow
scope reading for the subject quantifier) or with $\comu_3$ (going to the
component of the
object determiner and producing the
narrow scope reading for the object quantifier). The figure shows (in
the middle) the result of choosing $c_2-\comu_3$.

The term computed for component $C_2$ by following command $c_2$ is
$\cmdL{\lexseeks}{(x\backslash \beta)/y}$ and the $\comu_3$ link
joins components $C_2$ and $C_3$, replacing covariable $\alpha$ by
the complex context $\comu y. \cmdL{\lexseeks}{(x\backslash
  \beta)/y}$, producing the configuration shown schematically in the
middle of Figure~\ref{fig:ex_readingb} (refer back to
Figure~\ref{fig:semb} to see the initial labels). 

From this middle configuration and given the restriction to $\comu_2$
for the link connecting the two remaining components, only the command
$c_3$ is a possible in combination with the $\comu_2$ link. Command
$c_3$ would produce $\cmdL{\lexdet}{\alpha/\lexuni}$ from the
configuration shown in Figure~\ref{fig:semb} but given the previous
substitution for $\alpha$ it will now produce $\cmdL{\lexdet}{(\comu y. \cmdL{\lexseeks}{(x\backslash
  \beta)/y})/\lexuni}$ and the $\comu_2$ link will replace covariable
$\gamma$  by the context $\comu x. \cmdL{\lexdet}{(\comu y. \cmdL{\lexseeks}{(x\backslash
  \beta)/y})/\lexuni}$ and produce the configuration shown
schematically on the right of Figure~\ref{fig:ex_readingb}.

The final command $c_1$ produces $\cmdL{y'}{(\comu x. \cmdL{\lexdet}{(\comu y. \cmdL{\lexseeks}{(x\backslash
  \beta)/y})/\lexuni})/z'}$, the cotensor link the extended command
\begin{equation}\frac{y' z'}{\lexsuj}. \cmdL{y'}{(\comu x. \cmdL{\lexdet}{(\comu y. \cmdL{\lexseeks}{(x\backslash
  \beta)/y})/\lexuni})/z'}
\end{equation}
 and, finally, by $\mu_1$ the term for the complete proof net (and its command-$\mu/\comu$ pairing):
\begin{equation}\label{reading:sov}
\mu\gamma. \frac{y' z'}{\lexsuj}. \cmdL{y'}{(\comu x. \cmdL{\lexdet}{(\comu y. \cmdL{\lexseeks}{(x\backslash
  \beta)/y})/\lexuni})/z'}
\end{equation}
Similarly, starting with the $c_2-\comu_2$ pairing will produce.
\begin{equation}\label{reading:osv}
\mu \beta. \cmdL{\lexdet}{(\comu y. \frac{y' z'}{\lexsuj}. \cmdL{y'}{(\comu x. \cmdL{\lexseeks}{(x\backslash \beta)/y})/z'})/\lexuni}
\end{equation}
These are the only two readings available with positive bias for the
two atomic $np$ arguments of the transitive verb, and, as we have seen
before, this gives the right quantifier scope possibilities for an
extensional transitive verb such as ``likes'' we have seen in
equations~(\ref{positiveone}) and (\ref{positivetwo}) (apart from the
variable names, equation (\ref{reading:osv}) differs from
(\ref{positivetwo}) in that the extended command fraction in the latter
term is at the innermost position, but the terms are equivalent up to
commutative conversions). 

When we use a negative bias for the two $np$ arguments of the
transitive verb, we obtain the following term, corresponding to
equation~(\ref{negativeone}).

\begin{equation}\label{reading:int}
\mu\beta . \frac{y' z'}{\lexsuj}. \cmdL{y'}{(\comu x. \cmdL{\lexseeks}{(x\backslash \beta)/(\mu\alpha. \cmdL{\lexdet}{\alpha/\lexuni})})/z'}
\end{equation}

\editout{
For an intensional verb
like ``needs'',
the object $np$ is not assigned a lexical bias (an can therefore be negative) and this gives a third
possibility: $c_3-\mu_3$ will then be possible as a first step,
producing the third reading.
\begin{equation}\label{reading:int}
\mu\beta . \frac{y' z'}{\lexsuj}. \cmdL{y'}{(\comu
  x. \cmdL{\lexseeks}{(x\backslash
    \beta)/(\mu\alpha. \cmdL{\lexdet}{\alpha/\lexuni})})/z'}
\end{equation}}


\section{Conclusions}

The Lambek-Grishin calculus is a symmetric version of the Lambek
calculus. Together with the interaction principles, it allows for the treatment of
patterns beyond context-free which
cannot be satisfactorily handled in the Lambek calculus.
We have compared two proof systems for \LG: focused
sequent proofs and proof nets. Focused proofs avoid the spurious
non-determinism of backward-chaining search in the sequent calculus; they
provide a natural interface to semantic interpretation via their continuation-passing-style
translation. Proof nets present the essence of a derivation in a visually appealing
form; they do away with the syntactic clutter of sequent proofs,
and compute the structure of the end-sequent in a data-driven manner 
where this structure has to be given before one can a start backward-chaining
sequent derivation. Proof terms are read off from the composition graph
associated with a net. The computation of these terms depends both on a bijection between
premise and conclusion atomic formulas and between command and $\mu/\comu$ axioms. As a
result, one net can be associated with multiple construction recipes (proof terms),
corresponding to multiple derivations in the focused sequent calculus.


\tableofcontents

\bibliographystyle{chicago}
\bibliography{verybig}

\end{document}